\documentclass[10pt,twocolumn,letterpaper]{article}

\usepackage[pagenumbers]{wacv} % To force page numbers, e.g. for an arXiv version

\usepackage{graphicx}
\usepackage{amsmath}
\usepackage{amssymb}
\usepackage{booktabs}
\usepackage{multirow}
\usepackage{pifont}
\newcommand{\cm}{\ding{51}}
\newcommand{\xm}{\ding{55}}
\usepackage{tablefootnote}
\usepackage[table]{xcolor}
\usepackage{enumitem}
\usepackage{soul}
\usepackage{siunitx}
\usepackage{algorithm2e}
\usepackage{placeins}

\usepackage[pagebackref,breaklinks,colorlinks]{hyperref}

\usepackage[capitalize]{cleveref}
\crefname{section}{Sec.}{Secs.}
\Crefname{section}{Section}{Sections}
\Crefname{table}{Table}{Tables}
\crefname{table}{Tab.}{Tabs.}

\SetKwFor{For}{for }{ $\lbrace$}{$\rbrace$}
\SetKwComment{Comment}{/* }{ */}

\def\wacvPaperID{320} % *** Enter the WACV Paper ID here
\def\confName{WACV}
\def\confYear{2024}

\begin{document}

\title{IndustReal: A Dataset for Procedure Step Recognition Handling\\ Execution Errors in Egocentric Videos in an Industrial-Like Setting}

\author{Tim J. Schoonbeek$^1$, Tim Houben$^1$, Hans Onvlee$^2$, Peter H.N. de With$^1$, Fons van der Sommen$^1$\\
$^1$Eindhoven University of Technology, Netherlands \hspace{0.5cm} $^2$ASML Research, Netherlands \\
{\tt\small t.j.schoonbeek@tue.nl}}
\maketitle

%%%%%%%%% ABSTRACT
\begin{abstract}
   Although action recognition for procedural tasks has received notable attention, it has a fundamental flaw in that no measure of success for actions is provided. This limits the applicability of such systems especially within the industrial domain, since the outcome of procedural actions is often significantly more important than the mere execution. To address this limitation, we define the novel task of \emph{procedure~step~recognition} (PSR), focusing on recognizing the correct completion and order of procedural steps. Alongside the new task, we also present the multi-modal \emph{IndustReal} dataset. Unlike currently available datasets, IndustReal contains procedural errors (such as omissions) as well as execution errors. A significant part of these errors are exclusively present in the validation and test sets, making IndustReal suitable to evaluate robustness of algorithms to new, unseen mistakes. Additionally, to encourage reproducibility and allow for scalable approaches trained on synthetic data, the 3D~models of all parts are publicly available. Annotations and benchmark performance are provided for action recognition and assembly state detection, as well as the new PSR task. IndustReal, along with the code and model weights, is available at: \\
   {\tt\small \url{https://github.com/TimSchoonbeek/IndustReal}}.
\end{abstract}

\vspace{-0.5cm}
%%%%%%%%%%%%%%%%%%%%%%%%%%%%%%%%%%%%%%%%%%%%%%%%%%%%%%%%%%%%%%%
%%%                     INTRODUCTION                        %%%
%%%%%%%%%%%%%%%%%%%%%%%%%%%%%%%%%%%%%%%%%%%%%%%%%%%%%%%%%%%%%%%
\section{Introduction}
% What are procedural actions and why would it be valuable to be able to understand the actions properly.
Imagine an engineer who has just finished a service action on an internal combustion engine, only to discover that a step was missed early in the procedure. The service engineer has to undo most of the work completed after the mistake to rectify the forgotten step. The correct servicing of an engine is an example of a procedure, i.e., a given set of instructions, describing the procedural actions required to complete a task. An algorithm would become of significant value if it can understand procedural actions by automatically recognizing and tracking steps during the execution of a procedure. A system containing such an algorithm can warn users about potential mistakes or forgotten steps \cite{EgoProceL}, and summarize the execution of a procedure, thereby eliminating the need for manual logbook keeping~\cite{IEEEVR_paper}. Understanding procedures is not only highly relevant for industrial tasks, but to a broad range of procedural tasks, such as tracking the stages of surgeries to enhance post-surgical assessment and optimizing procedure workflow~\cite{garrow2021machine,zisimopoulos2018deepphase}.

% Why are they inherently difficult to understand.
Automated understanding of procedures is a difficult task in computer vision for various reasons. Firstly, there is often a limited visual difference between subsequent steps. For instance, determining whether a screw is correctly mounted into a specific component requires a fine-grained visual understanding. Additionally, there can be a high degree of symmetry and a low degree of textural variation for objects within procedures, especially for industrial actions~\cite{t-less}. Secondly, it is not feasible to collect a vast amount of data for many procedures, as they are often infrequently performed and rather specialized. Finally, procedures can typically be completed correctly via several possible execution orders. Therefore, it is frequently not sufficient to merely look whether a single, pre-defined step is completed correctly.

\begin{figure*}
\begin{center}
\includegraphics[width=0.99\linewidth]{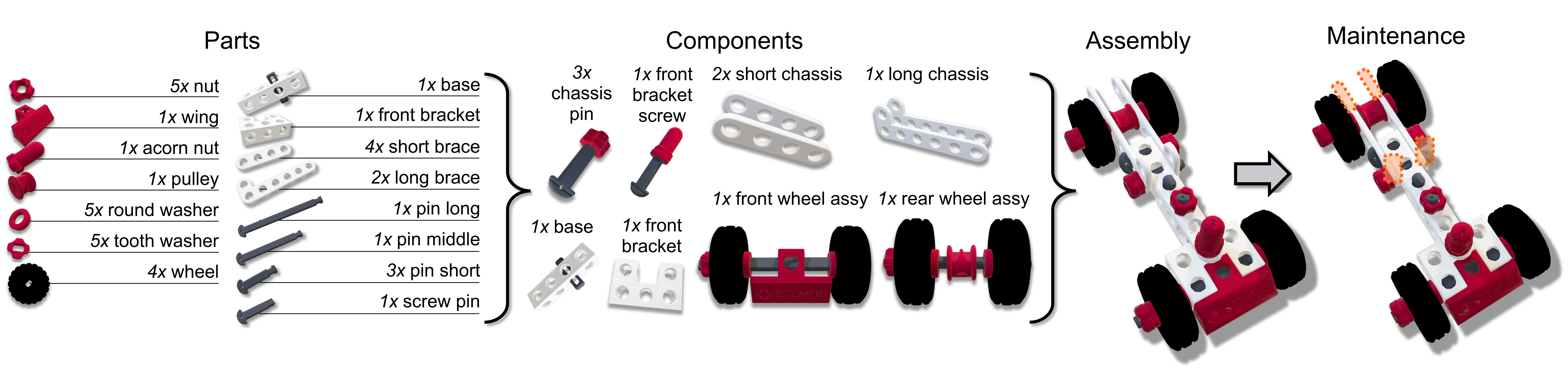}
\end{center}
\caption{Overview of the construction-toy car's (3D~printed) parts, components, and models used in IndustReal. The maintenance task represents a component upgrade and consists of replacing the long braces with short braces on the rear chassis.}
\label{fig: industrial_model}
\end{figure*}

% state of the art
Existing approaches to procedural understanding can generally be divided into two groups, one performing action recognition~(AR)~\cite{MEC,ASS} and the other assembly state detection~(ASD)~\cite{multistate_pe,Liu2020TGA,intelligent_ar_assmebly}. AR approaches aim to recognize only which actions are being performed, rather than which actions are actually completed. This is a crucial difference, and it is more valuable to know whether a step has been actually completed correctly, than to know only if an operator spend time on that step. ASD relies on object detection algorithms, detecting the actual phase during an object's assembly within a procedure. However, the number of possible states in ASD explodes with the number of parts in a procedure, limiting the complexity in existing literature to five or six parts only~\cite{multistate_pe,Liu2020TGA,intelligent_ar_assmebly}. Furthermore, both AR and ASD implementations do not explicitly leverage procedural knowledge, \eg which actions are to be expected after the observation of a preceding action. 
% Address dataset size problem, generalization to unseen errors, etc etc

To address the aforementioned limitations, this work formally defines the novel task of \textit{procedure step recognition}~(PSR) and introduces IndustReal, a publicly available dataset towards solving this task. It is an ego-centric, multi-modal dataset where 27~participants are challenged to perform assembly and maintenance procedures on a construction-toy car, based on STEMFIE~\cite{STEMFIE}, demonstrated in \cref{fig: industrial_model}. The videos are annotated for action recognition (AR), assembly state detection (ASD), and procedure step recognition. The placement of IndustReal within literature is outlined in \cref{tab: compare}. The IndustReal dataset features four novel aspects:
\begin{itemize}[topsep=0pt, itemsep=-4pt]
    \item \textit{Variety of execution errors.} IndustReal features 38~errors, of which 14~are exclusive to the validation and test sets. Whilst some datasets already include procedural errors (\eg omissions)~\cite{ASS,BRI}, IndustReal is the first to also include execution errors (\eg wrong type of nut used).
    \item \textit{Subgoal oriented execution.} Other datasets are either ``free-style'' assemblies~\cite{ASS} or contain a strict, step-by-step execution order~\cite{MEC}. IndustReal combines these execution types with a subgoal-oriented assembly style, where participants are given flexibility to determine the execution order between subgoals. This approach more closely resembles industrial procedures, since it maintains a hierarchy in procedure execution whilst allowing for flexibility where possible. IndustReal contains 48~different execution orders.
    \item \textit{Open-source geometries.} Scalability is an important factor for many industrial tasks, where simultaneously the technical drawings are often available. Therefore, 3D~models for all parts are published, to stimulate use of synthetic data in procedural action understanding, \eg by sim2real domain adaptation or generalization.
    \item \textit{3D~printed parts.} To ensure reproducibility, future availability of the model, and growth via community effort, all parts are 3D~printed and open source~\cite{STEMFIE}.
\end{itemize}
\medskip
In summary, the contribution of this work is two-fold: we present the IndustReal dataset, define the task of procedure step recognition and provide a benchmark towards this task.

\begin{table*}
  \begin{center}
    {\small{
\begin{tabular}{l|c|c|ccc|cccc|c|lrc}
\toprule
\multicolumn{1}{c|}{}        & \multicolumn{1}{c|}{} &     & \multicolumn{3}{c|}{Tasks} & \multicolumn{4}{c|}{Procedural complexity} &        & \multicolumn{3}{c}{Dataset size} \\
\multicolumn{1}{c|}{Dataset} & Year                  & Ego & AR      & ASD     & PSR     & Flex.      & PEs     & EEs     & Parts & 3DM    & Seqs.   & Dur.  & Participants  \\
\midrule
IKEA ASM~\cite{ASM}          & 2021                  & \xm & \cm     & \xm     & \xm     & \cm        & \xm     & \xm     & \hspace{0.2cm}7         & \xm    & 381    & 35.3h          & 48              \\
\rowcolor[gray]{.925}
MECCANO~\cite{MEC, MEC2}           & 2021                  & \cm & \cm     & \xm     & \xm     & \xm        & \cm$^*$     & \xm     & 49        & \xm    & 20     & 6.9h           & 20              \\
Assembly101~\cite{ASS}       & 2022                  & \cm & \cm     & \xm     & \xm     & \cm        & \cm     & \cm$^*$     & 15        & \xm    & 1.01K  & 167.0h         & 53              \\
\rowcolor[gray]{.925}
BRIO-TA~\cite{BRI}           & 2022                  & \xm & \cm     & \xm     & \xm     & \cm        & \cm     & \xm     & 10        & \xm    & 75     & 2.9h           & 15              \\
HA4M~\cite{HA4M}           & 2022                  & \xm & \cm     & \xm     & \xm     & \cm        & \xm     & \xm     & 17        & \cm    & 217     & 5.9h           & 41              \\
ATTACH~\cite{ATT}            & 2023                  & \xm & \cm     & \xm     & \xm     & \cm        & \xm     & \xm     & 26        & \xm    & 378    & 17.2h          & 42              \\
\midrule
\rowcolor[gray]{.925}
IndustReal (ours)            & 2024                  & \cm & \cm     & \cm     & \cm     & \cm        & \cm     & \cm     & 36        & \cm    & 84     & 5.8h          & 27            \\
\bottomrule
\end{tabular}
}}
\end{center}
\caption{Comparison of industrial-like procedural understanding datasets. AR: action recognition, ASD: assembly state detection, PSR: procedure step recognition, flex.: more than one single execution order for the task, PEs: procedural errors (omission, execution order), EEs: execution errors (component installed incorrectly), 3DM: 3D~models made publicly available, $^*$rare and not explicitly annotated.}
\label{tab: compare}
\end{table*}
%%%%%%%%%%%%%%%%%%%%%%%%%%%%%%%%%%%%%%%%%%%%%%%%%%%%%%%%%%%%%%%
%%%                     Related Work                        %%%
%%%%%%%%%%%%%%%%%%%%%%%%%%%%%%%%%%%%%%%%%%%%%%%%%%%%%%%%%%%%%%%
\section{Related Work}

\subsection{Work on procedural action understanding}
In (procedural) action recognition tasks, the objective is to classify a video clip into a set of activities expected within a certain procedure~\cite{multistate_pe,Liu2020TGA,intelligent_ar_assmebly,egot2}. Examples of such procedures are cooking~\cite{Damen2018EPICKITCHENS,zhukov2019cross,bansal2022my}, instruction videos~\cite{shen2021learning,lin2022learning}, and assembly tasks~\cite{han2017,epic_tent}. Recently, focus has been directed towards recognizing procedural actions in industrial-like settings~\cite{MEC,ASS,ATT,BRI,ASM}. 

Whilst recognizing actions in procedural videos is certainly of some interest in an industrial setting, we argue that it is more valuable to know what was actually completed, rather than only performed (and therefore potentially not finished). Therefore, the PSR task is proposed to recognize the correct completion of steps and to track the order in which those steps were completed.

Additionally, the comparable datasets outlined in \cref{tab: compare} generally have restricted sizes, since it is challenging to record sufficient video data of people executing (nearly) the same procedural task. This is especially relevant to industrial applications, where a significant portion of the tasks is performed rather infrequently. However, procedural knowledge, in the form of work instructions or technical drawings, is generally available for those tasks. Whilst such procedural knowledge is not explicitly leveraged by the aforementioned approaches, we advocate to explicitly use this knowledge in PSR to constrain the number of possible actions expected at any given moment.

\subsection{Literature on assembly state detection}
Assembly state detection is a sub-task of object detection, where the objective is to recognize and locate the specific state of an object during an assembly procedure~\cite{multistate_pe,Liu2020TGA,intelligent_ar_assmebly,ikea_state_dataset,yin2018synchronous,ar_visual_recognition}. ASD is significantly more challenging than object detection, since the visual variability between two states is often small. For instance, detecting the object ``Car'' is significantly less complicated than detecting the state ``Car without window''. Su~\textit{et~al.}~\cite{multistate_pe} simultaneously detect the state and the pose of a coffee machine during its assembly. The authors demonstrate a good performance on the task, but the different assembly states consist of visually distinctive objects. Lui~\textit{et~al.}~\cite{Liu2020TGA} demonstrate an attention mechanism for their convolutional neural network (CNN), detecting object states even if they are visually similar to each other. Both works demonstrate promising advances on this task. The approaches are trained predominantly on synthetic data, which can be readily generated if 3D~models of each assembly state are already designed.

Nevertheless, the aforementioned approaches have two important limitations. Firstly, the algorithms must be trained on a dataset of images from each object state. This forces the models to learn a low-dimensional representation for each state. Since the procedures that the authors selected consist of only 5~parts and, at most, 6~distinctive object states, training a model in such a manner is feasible. However, it remains unclear whether this approach scales to procedures with higher complexity and more object states. Secondly, no procedural information is leveraged by either approach. Therefore, the models expect for example, the initial state in an assembly equally as much as the very last state. For tasks with sufficient visual distinction between states, this is not a critical limitation. However, for objects that appear identical from certain viewpoints for several different states, procedural information could be used to determine the most likely state. Finally, none of the above-mentioned approaches release their (test) data.

\subsection{Published comparable datasets}
The placement of IndustReal within publicly available datasets, recorded in industrial-like settings is shown in \cref{tab: compare}. The MECCANO dataset~\cite{MEC, MEC2} is most relevant to IndustReal in terms of procedure, task complexity, and dataset size. Notable differences between these two datasets are that in MECCANO users follow strict, step-by-step instructions, limiting the variety in execution order. Secondly, although MECCANO contains some procedure errors, they are nearly always corrected in the subsequent step. Therefore, later states do not have prior errors in them, whilst such cases are certainly of interest and are frequently encountered in the industrial domain. Finally, we argue that the availability of 3D~models of all parts represented in the dataset is crucial for industrial applications, given the wide availability and usage of technical drawings in that domain. Unfortunately, the MECCANO dataset uses a IP-protected construction set, prohibiting the publication of such CAD models. BRIO-TA~\cite{BRI} and the large-scale Assembly101~\cite{ASS} datasets both contain procedural mistakes, such as omissions and incorrect execution order, but do not contain (labeled) execution errors. Furthermore, none of the mentioned datasets provide labels for assembly state detection. Lastly, to ensure future availability of the models, all parts in IndustReal are 3D~printed. An added benefit of 3D~printing is that it allows researchers to print in different scale, colors, or materials, to test the limits of their algorithms.

%%%%%%%%%%%%%%%%%%%%%%%%%%%%%%%%%%%%%%%%%%%%%%%%%%%%%%%%%%%%%%%
%%%                     Procedure Step Recognition         %%%
%%%%%%%%%%%%%%%%%%%%%%%%%%%%%%%%%%%%%%%%%%%%%%%%%%%%%%%%%%%%%%%
\section{Procedure Step Recognition}
The previous section has identified a gap in the related work between AR and ASD. By formalizing the task of procedure step recognition~(PSR), along with an evaluation scheme, we encourage researchers to develop methods to automatically recognize the completion of steps, rather than the (partial) execution of activities. Additionally, PSR systems should explicitly leverage procedural knowledge and allow a flexible execution order for procedural tasks, when the procedure allows it. 

\subsection{Task definition}
\label{sec: task_def}
The objective of PSR is to extract an estimate of all procedure steps correctly performed by a person up to time $t$, based on sensory inputs $X_t=(x_t, x_{t-1}, \dots, x_{t-h})$ and a descriptive set of the procedural actions to be performed $\mathcal{P}=\{a_0, a_1, \dots, a_n\}$. Here, $h$ is the observation horizon and $n+1$ the total number of actions~$a_i\in \mathcal{P}$ covered in the procedure. The predicted completed procedure steps $\hat{y}_t$ at time $t$, given some computational model~$\mathcal{F}$, are defined such that 
\begin{equation}
    \hat{y}_t = \mathcal{F} (X_t, \mathcal{P}).
\end{equation}
Here, a ``predicted'' (recognized) procedure step does not refer to the prediction of a future step, but rather the prediction by a model for a correctly completed step, based on the given inputs. Crucially, this definition allows for real-time operation, since contrary to existing tasks, PSR does not require a full recording of the procedure as input~\cite{lin2022learning,lea2016edit}.

Each unique action $a_i\in \mathcal{P}$ contains information regarding this specific action, and can be different for varying approaches to PSR. For instance, $a_i$ can contain a step description, a template image of what the completed step is supposed to look like, or relative positions between components. Sensory inputs $X_t$ may comprise of camera images, depth maps, or even user interaction with the system, during the execution of a procedure. Note that with the given definition, a PSR system can be an ensemble of various computer vision algorithms, \eg object detection to determine the assembly state and action recognition to determine the activities.

The predicted procedure steps form an ordered list of elements, describing the completed steps and is defined as \begin{equation}
    \hat{y}_t = (\hat{s}_{\sigma (0)}, \hat{s}_{\sigma (1)}, \dots, \hat{s}_{\sigma (m)}),
\end{equation}
where step~$\hat{s}_{\sigma (i)}$ is the predicted completion of the action~$a_i\in \mathcal{P}$, at prediction time~$\hat{t}_{\sigma (i)}$, having a prediction confidence~$c_{\sigma (i)}$, with $m+1$ the total number of recognized procedure steps. The function $\sigma$ maps a completed step~$\hat{s}_j\in \hat{y}_t$ to the corresponding action~$a_i\in \mathcal{P}$. Therefore, the first predicted step~$\hat{s}_{\sigma (0)}$ is not necessarily the completion of the action~$a_0\in \mathcal{P}$, but rather the first recognized step that a person completed.

The ground-truth execution order~$y_t$ at time~$t$ describes the order in which the steps are actually completed, which can differ from the prescribed order in $\mathcal{P}$, and is defined as
\begin{equation}
    y_t = (s_{\rho (0)}, s_{\rho (1)}, \dots, s_{\rho (k)}),
\end{equation}
where $s_{\rho (i)}$ is the completion of the action~$a_i\in \mathcal{P}$ at time~$t_{\rho (i)}$. The value $k+1$ is the total number of actions completed and $\rho$ a function that maps the completed steps from $y_t$ to the steps described in $\mathcal{P}$. If the order of the steps predicted in $\hat{y}_t$ equals that of $y_t$, it follows that $\sigma = \rho$, signifying a perfect execution order prediction.

\subsection{Evaluation metrics}
To quantify the performance of a PSR system, three evaluation metrics are proposed. These metrics focus on predicting the procedure steps in the correct order, the number of false predictions, and the timeliness of the predictions. 

\paragraph{Metric 1: procedure order similarity.} We propose to measure the quality of a predicted sequence order for an entire recording, $\hat{y}$ (e.g., `ACB'), by comparing it with a similarity measure with respect to the ground-truth $y$ (e.g., `ABC'). This is approached as a string similarity problem, a common problem in spelling error detection~\cite{bard2006spelling,rinartha2018comparative}, since words consist of a sequence of characters, where order and type of character matters. In temporal action segmentation, the Levenshtein~(Lev) distance, normalized over the length of the ground-truth sequence, is commonly used~\cite{lea2016edit}. We propose two changes to this metric, by (1) eliminating substitution from the edit distance, preventing the metric from favouring models with many false positives, and (2) using the Damerau-Levenshtein~(DamLev)~\cite{damerau:1964:damlevdistance}, rather than the Lev distance because it penalizes transpositions less, since it is intuitive to penalize ``A\underline{CB}'' less compared to ``\underline{C}A\underline{B}''.

In contrast to Lea \etal~\cite{lea2016edit}, we propose to normalize the edit distance with respect to the length of the ground truth, rather than the length of either the ground truth or the prediction, depending on which is longer. This prevents models with many false positives from being normalized favourably. Finally, the normalized edit distance is subtracted from the unity value, resulting in a similarity metric, rather than a distance metric. Thus, the procedure order similarity~(POS) between $y$ and $\hat{y}$ is defined as
\begin{equation}
    \textrm{POS} = 1 - \textrm{min}(\frac{\textrm{DamLev}(y, \hat{y})}{|y|},\hspace{0.1cm}1),
\end{equation}
where $\textrm{DamLev}(\cdot)$ is a weighted DamLev edit distance function. Further clarification on POS may be found in~\cite{supp}.

\paragraph{Metric 2: $\mathbf{F_1}$ score.}
A false positive is defined as a procedure step~$\hat{s}_{\sigma (j)}$ that is predicted prior to the actual completion of action~$a_i$, or if $a_i$ is not at all completed, hence
\begin{equation}
    (\hat{t}_{\sigma (j)} < t_{\rho (i)}) \vee (a_i \not \in y).
    \label{eq: FP}
\end{equation}
A false negative is defined as a step~$s_{\rho (j)}$ for a corresponding action~$a_i$, that has indeed been completed, but is not represented in $\hat{y}$, so that
\begin{equation}
    (a_i \in y) \wedge (a_i \not \in \hat{y}).
    \label{eq: FN}
\end{equation}
Finally, a true positive is defined as the prediction of a procedure step~$\hat{s}_{\sigma (j)}$, that is observed at or after the actual completion of $a_i$, such that
\begin{equation}
    (\hat{t}_{\sigma (j)} \geq t_{\rho (i)}) \wedge (a_i \in y).
    \label{eq: TP}
\end{equation}

\begin{figure*}
\begin{center}
% \fbox{\rule{0pt}{4in} \rule{.9\linewidth}{0pt}}
\includegraphics[width=0.99\linewidth]{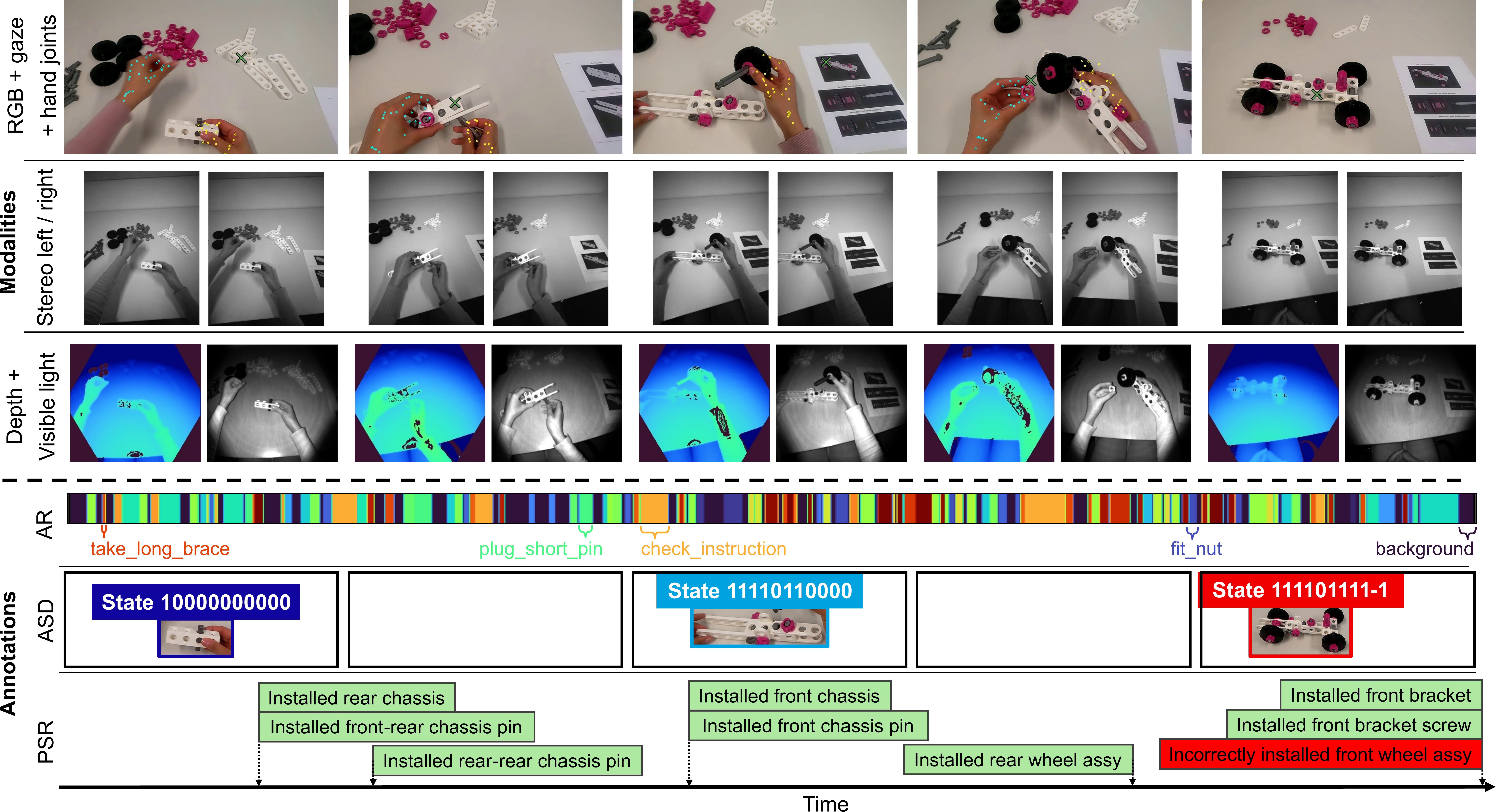}
\end{center}
  \caption{Samples from a clip in the IndustReal dataset, demonstrating the modalities and annotations for all three tasks. Gaze is indicated by the cross, detected hand joints by the dots. AR: action recognition, ASD: assembly state detection, PSR: procedure step recognition.}
\label{fig: dataset_sample}
\end{figure*}

PSR systems can explicitly assume that actions have been completed based on procedural information, without relying exclusively on sensory recognition. Therefore, a distinction can be made between $F_1$~score at the recognition and system level. Since this definition does not contain any time restriction on true positives, procedure steps that are recognized long after the step completion are not penalized. Therefore, comparing only the procedure order similarity and $F_1$~score lacks a temporal component.

\paragraph{Metric 3: average delay.}
To complement the aforementioned metrics with a temporal component, the average delay~$\tau$ is introduced, quantifying the time between the ground-truth completion and corresponding recognition of a step. False negatives, defined by \autoref{eq: FN}, have an undefined delay, and similarly, false positives (defined in \autoref{eq: FP}) have either an undefined delay, or a negative delay. Therefore, FPs and FNs are excluded from $\tau$, defines as
\begin{equation}
    \tau = \frac{1}{h} \sum_{i=0}^{h-1} (\hat{t}_{\sigma (i)} - t_{\rho (i)}), 
\end{equation}
where $h$ is the number of total TPs in $y$. Because FPs and FNs are discarded in the delay, only the combination of the three proposed metrics provides a valuable insight into the performance of a system towards solving PSR.

%%%%%%%%%%%%%%%%%%%%%%%%%%%%%%%%%%%%%%%%%%%%%%%%%%%%%%%%%%%%%%%
%%%                     IndustReal Dataset                  %%%
%%%%%%%%%%%%%%%%%%%%%%%%%%%%%%%%%%%%%%%%%%%%%%%%%%%%%%%%%%%%%%%
\section{IndustReal Dataset}
% This section describes the objects, procedures and annotations of the IndustReal dataset. 

\subsection{Construction-set car procedures}
\Cref{fig: industrial_model} demonstrates the 36~part models used in the IndustReal dataset, based on the STEMFIE construction-toy car~\cite{STEMFIE}. The model has significant complexity, consisting of multiple types of washers, pins, and braces. Some components require screwing, others need tightening, and participants frequently have to use both hands. Two procedures are defined, an assembly task, where the car has to be build from scratch, and a maintenance task, where the participants have to replace part of the rear chassis of the toy car. Printed instructions are provided in a subgoal-oriented manner, meaning that the participant builds towards subgoals rather than executing strict, step-by-step instructions, or ``free-style'' building towards a final assembly. Unlike related datasets, participants are allowed to create subassemblies, as commonly encountered in industrial settings. 

All of the parts are 3D~printed on an Ultimaker S5 at 200\% scale, layer height of 0.3~mm, 15\% infill, a print speed of 50~mm/s, and PLA filament and the colors white, silver metallic, magenta and black. All parts, as well as the final models, are published together with the dataset, since part geometries are commonly available in industrial settings, \eg CAD models.

\subsection{Recording and setting}
\label{sec: rec_setting}
The HoloLens~2~(HL2)~\cite{hl2} mixed reality headset is used as recording rig for the dataset. The front-facing RGB camera records at a resolution of 1280$\times$720 pixels and the stereo cameras provide images at 480$\times$620 pixels. The long-throw depth and IR sensors record at a resolution of 320$\times$288 pixels and contain normalized values. Next to the images, we also record gaze, hand, and head-pose tracking, provided by the HL2 algorithms. All sensors, visualized in \cref{fig: dataset_sample}, are sampled at 10~fps, except from the depth and IR sensors, as these are limited to 5~fps in hardware. The data are sent from the HL2 to a server in real time, using the HL2SS library~\cite{HL2SS}.

The dataset is recorded in a setting with consistent background and lighting conditions. More specifically, participants are asked to perform the procedures on a white desk placed against a white background. Such lack of variety in background and lighting conditions is assumed to be consistent with industrial settings. Redundant washers, nuts, screws, and pins are placed amongst the required parts to reduce bias towards detecting unused parts. Prior to each recording, all parts are sorted into heaps according to color, and then placed randomly on the desk.

\subsection{Participants and protocol}
In total, 27~participants were recruited. Each participant signed a consent form and the experiment was approved by the institution's Ethical Review Board. Each participant was asked to perform the assembly procedure once, without recording, to familiarize themselves with the construction-set car. During this practice assembly, feedback was provided to the participants when required. Subsequently, the HL2 was fitted to the participant and the gaze tracking was calibrated. Then, each participant was assigned a correct assembly instruction, a correct maintenance instruction, and one or two instructions with errors, to ensure that a variety of mistakes were introduced. As anticipated, the participants exhibited errors even when they were provided with the correct instructions, and these mistakes were annotated as well. The recorded errors in the dataset vary from minor, difficult-to-observe mistakes (\eg installing the wheels without washers), to large mistakes, such as forgetting a wheel. To minimize ``learners bias'' of the participants in the dataset, each participant was given the instructions in a random order. Participants were asked to remove their hands from the assembly after completing a step, to ensure a minimally occluded view on the assembly state.

\subsection{Annotation}
Since IndustReal is intended to be closely representative of industrial use cases, significant importance is given to evaluation and robustness to unseen, out-of-distribution errors. Additionally, a large real-world test set is desired when training exclusively on synthetic data. Therefore, a dataset split of 12/5/10 (No. of train/val/test participants) is chosen, which is heavily focused on the test set. Furthermore, numerous errors are exclusively present in the validation or test sets. Note that the split is made on participants, rather than videos, to ensure sufficient variation in viewpoint, execution, and head movements between the three sets. 

Although IndustReal is specifically presented to address PSR, annotations are also provided for AR and ASD, enabling researchers to use IndustReal for various tasks, or to combine tasks for better PSR performance in future work.

\subsubsection{Action recognition}
AR labels consist of the frame at which the action starts, ends, and the combination of a verb and noun. Given the resemblance between MECCANO~\cite{MEC} and the construction-toy car used in IndustReal, we adopt the same verbs as utilized in MECCANO: \textit{take}, \textit{put}, \textit{align}, \textit{plug}, \textit{pull}, \textit{screw}, \textit{unscrew}, \textit{tighten}, \textit{loosen}, \textit{fit}, \textit{check}, and \textit{browse}. Using these verbs, the component names described in \cref{fig: industrial_model}, and the additional nouns \textit{objects}, \textit{partial model}, and \textit{instruction}, we have annotated 75~fine-grained action classes and a total of 9,273~instances. The average action lasts 1.9$\pm$1.4~seconds. A long-tail distribution is observed, with 80\% of the data containing 29.3\% of the actions, and provided together with more statistical details in~\cite{supp}. Recordings have an average of 110$\pm$38~actions per video, with 134$\pm$32~actions per assembly and 79$\pm$13~actions maintenance video. Participants frequently perform actions simultaneously, resulting in 24.2\% of all instances having an overlap with at least one other action.

\subsubsection{Assembly state detection}
\label{sec:ASD}
To the best of our knowledge, IndustReal is the first publicly available dataset with ASD annotations. Because the participants have flexibility in execution order, all labeled states must be explicitly defined. We label the assembly states with integers, where a ``1'' indicates that a component at that index has been correctly installed and a ``0'' indicates that it has not (yet) been correctly installed. Additionally, we assign ``-1'' to components that have been incorrectly installed. Rather than labeling each assembly part individually with such a code, we divide the toy car into 11 components (in order): \textit{base}, \textit{front chassis}, \textit{front chassis pin}, \textit{rear chassis}, \textit{short-rear chassis}, \textit{front-rear chassis pin}, \textit{rear-rear chassis pin}, \textit{front bracket}, \textit{front bracket screw}, \textit{front wheel assy}, and \textit{rear wheel assy} (see \cref{fig: industrial_model}). For example, the assembly state \textit{11100000000} consists of a correctly installed base, front chassis and front chassis pin. 

Bounding boxes and labels are provided for all 22~labeled (defined) states in IndustReal, as well as 27~different error states. Intermediate states, which occur during the assembly of components, are not labeled, as outlined in \cref{fig: dataset_sample}. Whilst those states could be annotated as \textit{partial~model}, such labels empirically appear to hold little value. Intermediate states differ from erroneous states in that participants in intermediate states are actively progressing towards completing a state, whereas participants in erroneous states are not trying to further complete that step. In total, 26.9K~video frames (13\% of total) are annotated for ASD, of which 3,569~frames show error states.

\subsubsection{Procedure step recognition}
The PSR labels consist of the frame at which a step completion occurs and the new assembly state, as defined in the previous section. The difference between two assembly states can directly be used to determine which procedure steps are completed. A procedure step is defined as \textit{completed} when the component relating to that step is correctly installed, which includes actions such as the tightening of a nut. Although the explicit detection of incorrect step execution is not included in the PSR task definition (\cref{sec: task_def}), annotations of incorrect assembly states are included for qualitative analysis and future work. Therefore, two sets of PSR labels are provided, one with only correctly executed procedure steps (\eg ``Installed front chassis''), and one that also includes incorrectly completed steps, such as ``Incorrectly installed front wing''. An example of a PSR-annotated sequence is outlined in \cref{fig: dataset_sample}. In total, IndustReal consists of 724~correct procedure step completions (8.6$\pm$1.2~correct completions per recording) and 38~incorrect step completions. In total, 35~videos (42\%) contain a missing or incorrectly completed procedure step. IndustReal contains 22~different correctly completed procedure execution orders, plus an additional 26~different execution orders containing error states.

%%%%%%%%%%%%%%%%%%%%%%%%%%%%%%%%%%%%%%%%%%%%%%%%%%%%%%%%%%%%%%%
%%%                     Benchmark Experiments               %%%
%%%%%%%%%%%%%%%%%%%%%%%%%%%%%%%%%%%%%%%%%%%%%%%%%%%%%%%%%%%%%%%
\section{Benchmark Experiments}
This section outlines the benchmark performance of state-of-the-art approaches to AR and ASD on IndustReal. Furthermore, a PSR benchmark implementation is outlined and evaluated, providing a baseline performance for more sophisticated approaches towards this task.

\subsection{Action recognition benchmark}

\begin{table}
  \begin{center}
    {\small{
\begin{tabular}{ll|cc}
\toprule
Model                       & Modalities& \begin{tabular}[c]{@{}c@{}}Top-1\\ acc. {[}\%{]}\end{tabular} & \begin{tabular}[c]{@{}c@{}}Top-5\\ acc. {[}\%{]}\end{tabular} \\
\midrule
SlowFast~\cite{slowfast}$^*$          & RGB              & 57.83             & 82.87 \\
SlowFast~\cite{slowfast}$^\dagger$    & RGB              & 60.39             & 85.21 \\
MViTv2~\cite{mvitv2}$^*$              & RGB              & 62.43             & 85.62 \\
MViTv2~\cite{mvitv2}$^\dagger$        & RGB              & 65.25             & 87.93 \\
\midrule
SlowFast~\cite{slowfast}$^\dagger$    & RGB, VL, stereo  & 62.34             & 85.97 \\
MViTv2~\cite{mvitv2}$^\dagger$        & RGB, VL, stereo  & \textbf{66.45}    & \textbf{88.43} \\
\bottomrule
\end{tabular}
}}
\end{center}
\caption{AR benchmark on IndustReal. $^*$MECCANO~\cite{MEC} pre-trained, $^\dagger$Kinetics~\cite{kinetics-400} pre-trained, VL:~visible light.}
\label{tab: ar_bench}
\end{table}

\paragraph{Definition.} Given a video segment $X_i=[x_{ts_i}, x_{te_i}]$ and a set of action classes $C_a = \{c_0, c_1, ..., c_n\}$, the objective of action recognition is to classify segment~$X_i$ to the correct class~$c_i\in C_a$~\cite{MEC}. Here, $x_{ts_i}$ and $x_{te_i}$ indicate the start and end frame for action~$c_i\in C_a$, respectively.

\vspace{-0.5cm}
\paragraph{Benchmark.} For this task, the SlowFast~\cite{slowfast} CNN and MViTv2-S~\cite{mvitv2} transformer are chosen to benchmark the performance. Each model is trained on IndustReal after Kinetics~\cite{kinetics-400} pre-training. Additionally, since the MECCANO dataset for action recognition is closely related to the IndustReal dataset, we also report baselines pre-trained on MECCANO. Finally, both networks are trained on depth, visible light (VL), and stereo images, and combined to create an ensemble of models trained on various modalities.

Top-1 and top-5 accuracy are reported in \cref{tab: ar_bench} for the aforementioned experiments. Pre-training on the MECCANO dataset does not provide a performance benefit. The MViTv2 transformer outperforms the SlowFast architecture. The best performance is observed for the MViTv2 ensemble of the modalities RGB, VL, and stereo images. As motivated in~\cite{supp}, depth is excluded from this ensemble. Notably, for both architectures, each individual modality is outperformed by the ensemble, indicating that each modality contains some complementary form of relevant information.

\subsection{Assembly state detection benchmark}
\begin{table}
  \begin{center}
    {\small{
\begin{tabular}{ll|cc}
\toprule
Pre-trained          & Fine-tuned                                                        & \begin{tabular}[c]{@{}c@{}}mAP\\ (b-boxed)\end{tabular}  &  \begin{tabular}[c]{@{}c@{}}mAP\\ (entire videos)\end{tabular} \\
\midrule
COCO                 & Synthetic                                                         & 0.573                 & 0.341 \\
% COCO                 & IndustReal A                                                      & 0.788                 & 0.362 \\
COCO                 & IndustReal                                                        & 0.753                 & 0.553 \\
% Synthetic            & IndustReal A                                                      & 0.820                 & 0.367 \\
Synthetic            & IndustReal                                                        & 0.779                 & 0.575 \\
% COCO                 & \begin{tabular}[c]{@{}l@{}}IndustReal A + \\ synthetic\end{tabular} & \textbf{0.857}      & 0.637 \\
COCO                 & \begin{tabular}[c]{@{}l@{}}IndustReal   + \\ synthetic\end{tabular} & \textbf{0.838}               & \textbf{0.641} \\
\bottomrule
\end{tabular}
}}
\end{center}
\caption{ASD performance benchmark on IndustReal using YOLOv8-m~\cite{YOLOv8} for various training schemes.}
\label{tab: asd_bench}
\end{table}

\paragraph{Definition.} Given a video frame~$X_i$ and a set of assembly states $Z_a = \{z_0, z_1, ..., z_n\}$, the objective of ASD is to detect the bounding box and assembly state~$z_i\in Z_a$ for the sample~$X_i$. The states~$Z_a$ are defined in \cref{sec:ASD}.

\vspace{-0.5cm}
\paragraph{Benchmark.}
For this benchmark, the state-of-the-art object detection network YOLOv8-m~\cite{YOLOv8} is employed. Since geometries of all parts are published with IndustReal to stimulate synthetic learning, we provide four training schemes. First, we train the model, pre-trained on COCO~\cite{COCO}, exclusively on a synthetically generated dataset. To generate the synthetic data, Unity Perception~\cite{unity_perception} is used to generate 100K training samples, each containing one assembly state in $Z_a$. Secondly, we train the model (pre-trained on COCO) directly on IndustReal. Then, we pre-train on the synthetic dataset, after which we fine-tune on the IndustReal dataset. Finally, the synthetic and real-world datasets are combined for the last baseline.

The results are quantified using the mAP metric and reported on frames of the IndustReal test set containing ground-truth bounding-box annotations, as well as the entire test set. As outlined in \cref{tab: asd_bench}, combining the synthetic and real-world images results in the highest performance. A significant performance drop of 27\% is observed when evaluating entire videos, rather than only frames with a ground-truth annotation. The decreased performance is caused by false positive predictions on states with fine-grained visual differences, predominantly error states and directly prior to the completion of a procedural step. Notably, the best performing model has a false positive rate of 65\% and average precision (AP) of 0.23 for assembly states containing an error. Two such error states, and their corresponding ASD predictions, are visualized in \cref{fig: psr}, and further qualitative and quantitative analysis is provided in~\cite{supp}.

\subsection{Procedure step recognition benchmark}
\begin{table}[]
  \begin{center}
    {\small{
    \begin{tabular}{l|ccc|ccc}
    \toprule
     & \multicolumn{3}{c|}{All recordings} & \multicolumn{3}{c}{Recordings with errors} \\
              & POS   & $F_1$ & $\tau$ [s]  & POS   & $F_1$ & $\tau$ [s] \\
    \midrule
    B1        & 0.570 & 0.779 & \textbf{14.9}& 0.480 & 0.698 & \textbf{14.4} \\
    B1-S       & 0.014 & 0.206 & 36.9        & 0.000 & 0.174 & 48.4 \\
    B2        & 0.731 & 0.860 & 22.3        & 0.636 & 0.784 & 20.2 \\
    B2-S       & 0.240 & 0.573 & 44.4        & 0.107 & 0.516 & 60.5 \\
    B3        & \textbf{0.797} & \textbf{0.883} & 22.4 & \textbf{0.731} & \textbf{0.816} & 20.4 \\
    B3-S       & 0.597 & 0.734 & 49.5        & 0.571 & 0.731 & 71.4 \\
    \bottomrule
    \end{tabular}
    }}
    \end{center}
    \caption{PSR performance benchmark on IndustReal with ASD backbone, expressed by procedure order similarity (POS), $F_1$ score, and average delay $\tau$. B1-3 denote the three baseline models and S indicates exclusive training on synthetic data.}
    \label{tab: psr_bench}
\end{table}

\paragraph{Definition.} The task definition is given in \cref{sec: task_def}. 

\vspace{-0.5cm}
\paragraph{Benchmark.} For the PSR benchmark, we provide three baseline implementations relying on the ASD model output outlined in the previous section. The first baseline~B1 determines for each change in detected assembly state which corresponding steps must have been completed since the last detected state, and assumes these actions to be executed correctly. The second baseline~B2 accumulates the confidence for each predicted step completion (obtained from B1) over time, until a threshold~$T$ is reached, upon which an action is deemed correctly completed. Baseline 3 uses the same confidence aggregation as B2, but limits the number of possible step completions to those expected in the correct execution of the given procedure. Further clarification on the baselines may be found in~\cite{supp}. For each baseline, the best performing ASD model from \cref{tab: asd_bench} is used. Additionally, each baseline is evaluated with the ASD approach trained exclusively on synthetic data. The detection pipeline (ASD + PSR) yields real-time operation at 178~fps on a v100 GPU.

\begin{figure}[t]
  \centering
  \begin{subfigure}{0.48\linewidth}
    \includegraphics[width=0.99\linewidth]{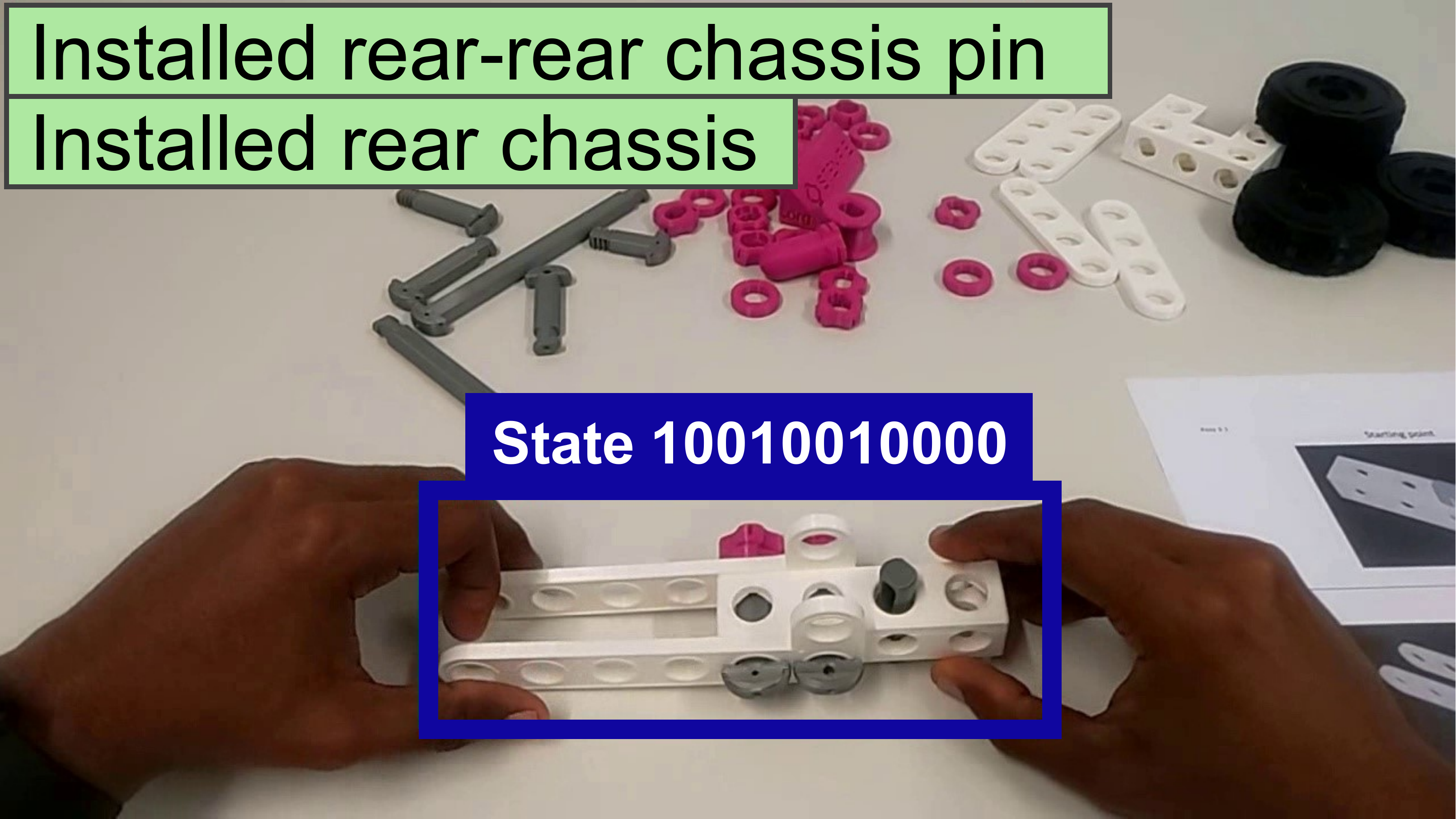}
    \caption{ASD predict: base, rear-rear pin and rear chassis correctly installed.}
    % \caption{Correct recognition of rear chassis and rear-rear pin. The incorrect front-rear pin orientation does not trigger a false positive recognition.}
    \label{fig: wrong_pin_ori}
  \end{subfigure}
  \hfill
  \begin{subfigure}{0.48\linewidth}
    \includegraphics[width=0.99\linewidth]{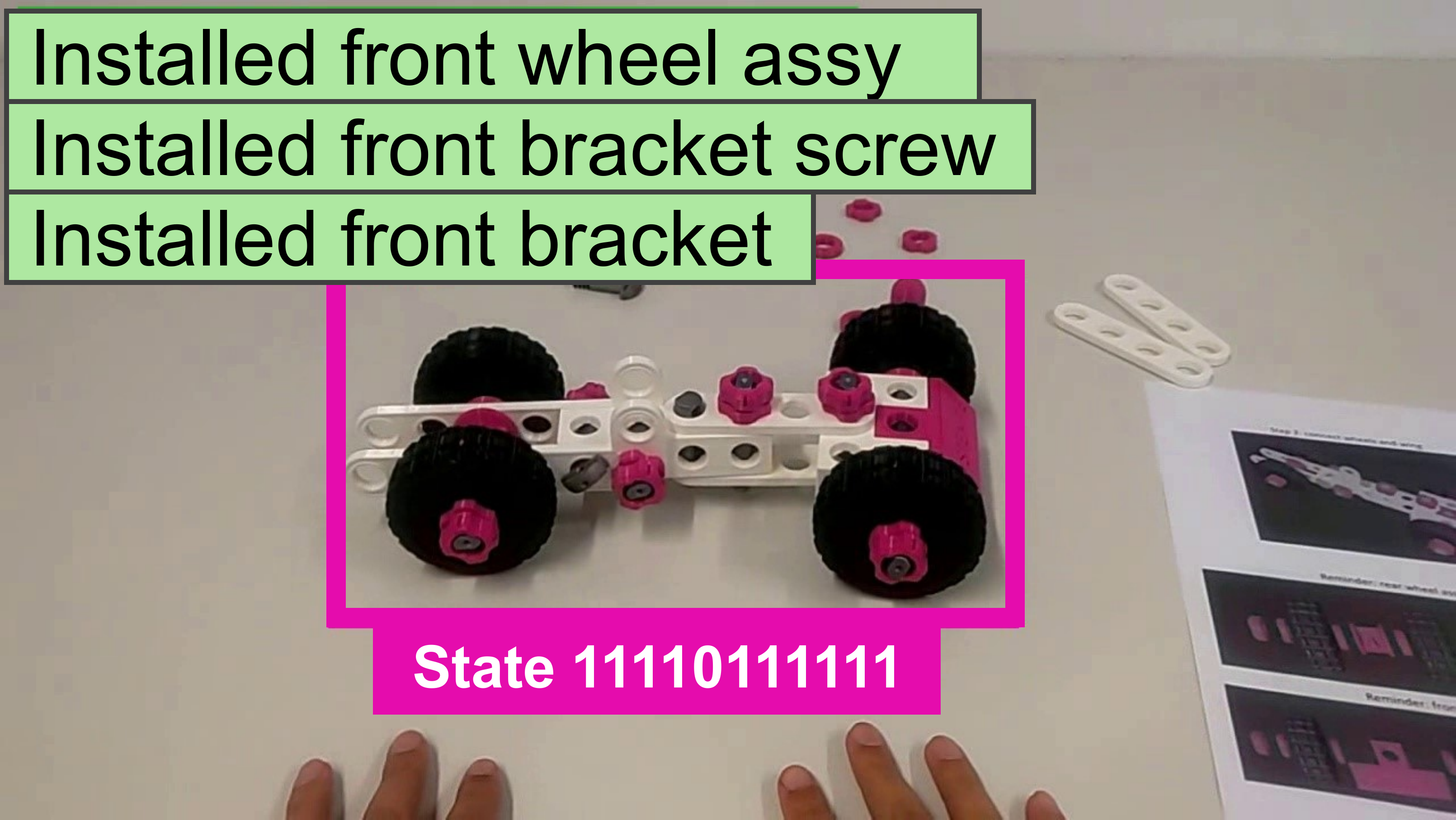}
    \caption{ASD predict: entire assembly procedure correctly installed.}
    % \caption{Recognition of correct front and wheel assy completion, but the incorrect use of a nut instead of screw for the front brace is not recognized.}
    \label{fig: wrong_nut}
  \end{subfigure}
  \vspace{0.25cm}
  \caption{Visualization of ASD and PSR results for two cases. Bounding boxes and captions indicate the ASD predictions, text boxes indicate PSR predictions (B3). Both ASD predictions are false positives. (a) Front-rear pin orientation is incorrect, whilst the ASD predicts the absence of this pin. (b) Nut instead of screw is incorrectly used for the front brace, which is not recognized.}
  \label{fig: psr}
\end{figure}

The POS, $F_1$, and $\tau$ metrics for all baselines are outlined in \cref{tab: psr_bench}. The best performing approach achieves relatively high POS and $F_1$ scores (0.797 and 0.883 on all recordings, respectively), due to under-representation of execution errors compared to correctly executed steps. On recordings with errors, all baselines show a distinctive decrease in performance, as shown in \cref{fig: psr}, thereby highlighting the need for approaches that better handle (out-of-distribution) errors. Although the performance of B3-S trained entirely on synthetic data is significantly lower than its real-world trained counterpart, it notably outperforms B2-S. This indicates that using procedural information to restrict the set of possible step completions can considerably improve the model performance.

%%%%%%%%%%%%%%%%%%%%%%%%%%%%%%%%%%%%%%%%%%%%%%%%%%%%%%%%%%%%%%%
%%%                     Conclusion                          %%%
%%%%%%%%%%%%%%%%%%%%%%%%%%%%%%%%%%%%%%%%%%%%%%%%%%%%%%%%%%%%%%%
\section{Conclusion}
This paper proposes the novel task of \emph{procedure step recognition}~(PSR). This task bridges the gap that currently existing \emph{action recognition} and \emph{assembly state detection} tasks leave in procedural activity understanding, by explicitly leveraging procedural information and recognizing correctly completed steps. This paves the way for the development of upcoming, increasingly powerful procedure-assistive systems.

Along with the PSR task, the IndustReal dataset is presented. IndustReal differs from existing datasets in the wide variety of procedural and execution errors, subgoal-oriented procedure execution, and 3D~printed parts to ensure future reproducibility. Additionally, the geometries of object parts are published to stimulate sim2real domain adaptation and generalization, as this is crucial to scalability in industrial settings. Furthermore, performance benchmarks are provided for AR, ASD, and PSR based on the proposed IndustReal dataset with various pre-training techniques.

The PSR baseline shows promising results on procedural videos where participants do not make mistakes, but fails to generalize to (out-of-distribution) execution errors. Future work will focus on improving upon this baseline performance and increase scalability, such that PSR becomes viable and indeed attractive towards more industrial use cases.

\section*{Acknowledgment}
The authors sincerely express their gratitude to Dr. Jacek Kustra for his valuable feedback and Goutham Balachandran for his help with annotations. Additionally, we thank all participants for their contribution. This work is partially executed at ASML Research and has received funding from ASML and the TKI research grant (project number TKI2112P07).

\section{\textit{Supplementary Material for:} \newline IndustReal: A Dataset for Procedure Step Recognition Handling Execution Errors in Egocentric Videos in an Industrial-Like Setting}

\subsection{Introduction}
This supplementary material accompanies the main paper titled \textit{IndustReal: A Dataset for Procedure Step Recognition Handling Execution Errors in Egocentric Videos in an Industrial-Like Setting}. This document offers readers a comprehensive and more in-depth understanding of the conducted research.

Section \ref{sec: AR} provides additional information on the action recognition (AR), such as the distribution of class instances, implementation details, and further quantitative results. Section \ref{sec: ASD} provides implementation details on assembly state detection (ASD) and outlines additional qualitative as well as quantitative results. Lastly, Section \ref{sec: PSR} provides further clarification on the motivation for the proposed procedure order similarity (POS) evaluation metric, implementation details and pseudo code describing the procedure step recognition (PSR) baselines, and a qualitative example.

\begin{figure}[t]
    \centering
    \includegraphics[width=0.9\linewidth,trim={35cm 2cm 19cm 7cm},clip]{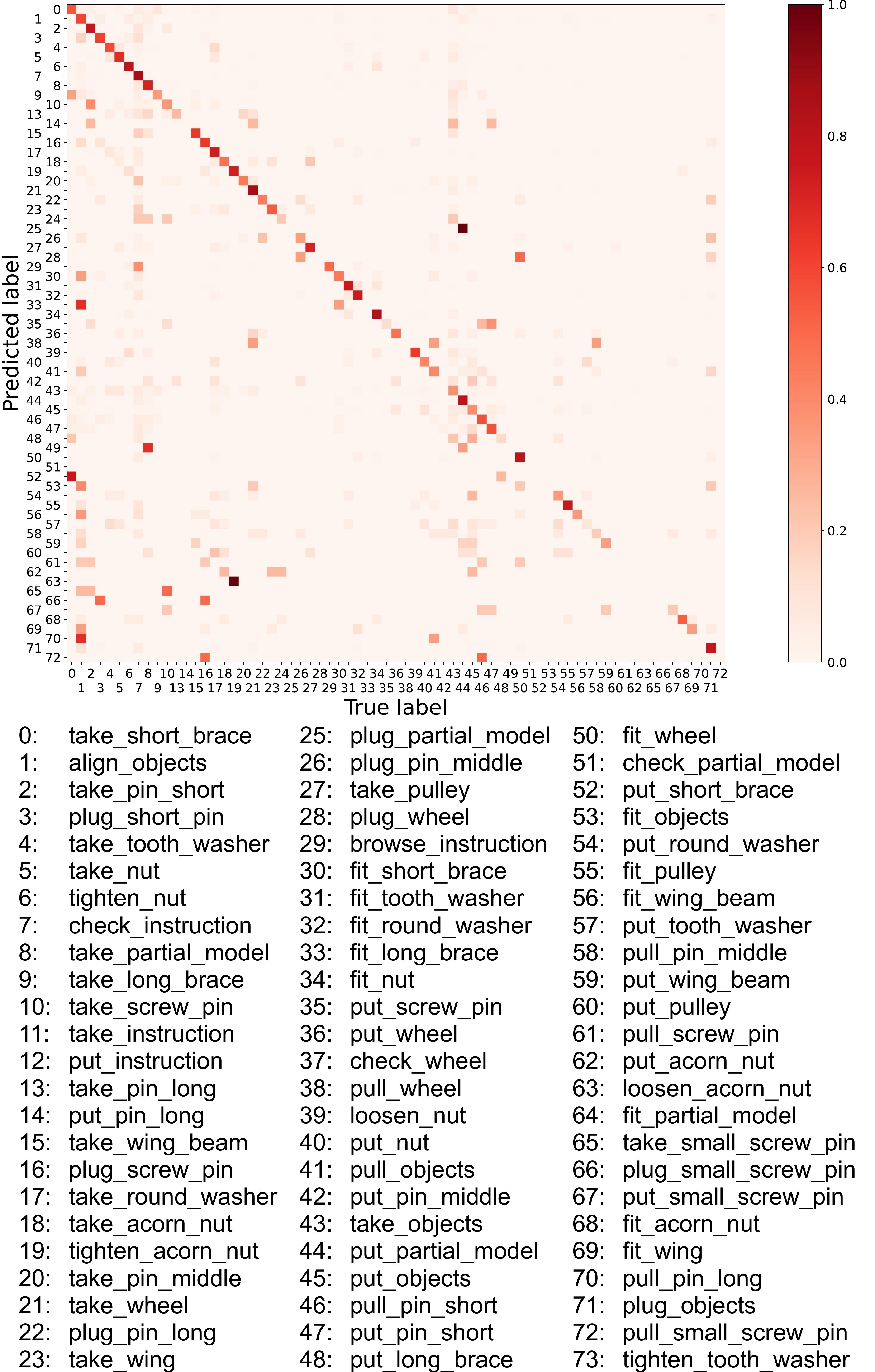}
    \caption{Normalized confusion matrix for action recognition (MViTv2~\cite{mvitv2} pretrained on Kinetics~\cite{kinetics-400}).}
    \label{fig: ar_conf}
\end{figure}

\begin{figure*}[]
    \centering
    \includegraphics[width=0.99\linewidth]{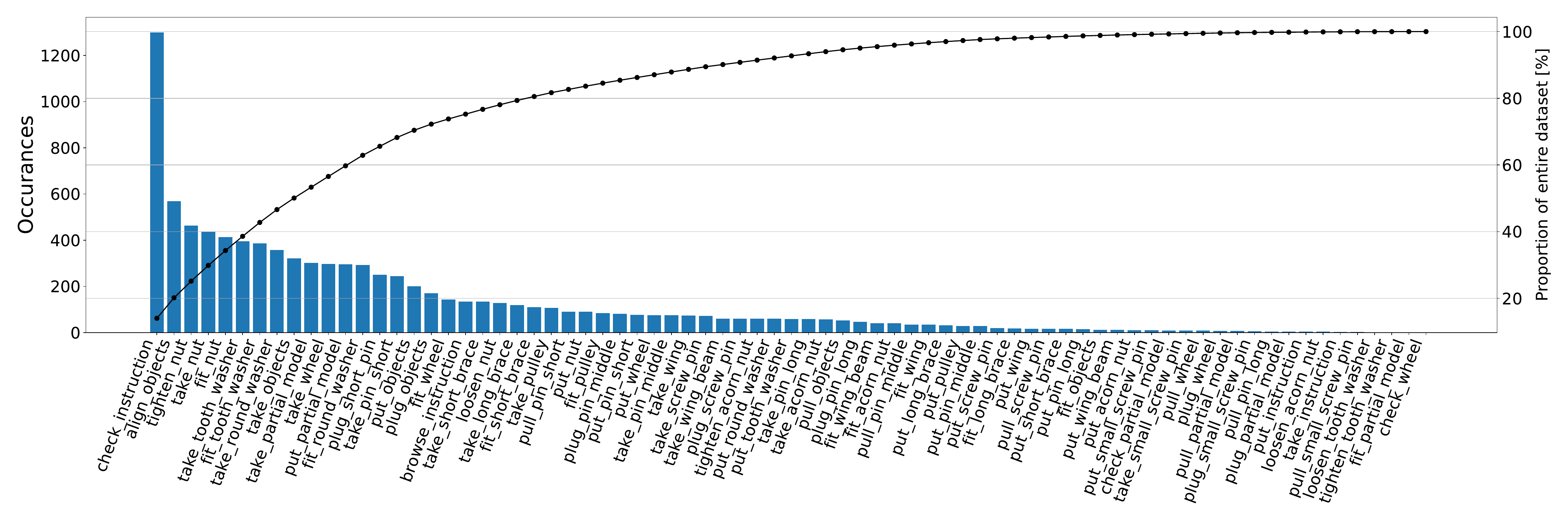}
    \caption{Visualization of the distribution of action recognition labels in the IndustReal dataset.}
    \label{fig: ar_labels}
\end{figure*}

\subsection{Action recognition}
\label{sec: AR}
\subsubsection{Annotations}
Annotators were instructed to mark the action start when the participants initiate the action, rather than when participants touch the relevant object(s). Figure~\ref{fig: ar_labels} demonstrates the action class instances and long-tail distribution for action recognition annotations in the IndustReal dataset. As can be seen, the \textit{check\_instruction} action class is the most common, followed by \textit{align\_objects}. The long-tail distribution demonstrates that 23 action classes (out of 73) constitute 80\% of the dataset. All actions were annotated using the ELAN tool~\cite{ELAN}. Two annotators annotated the entire dataset, and each annotator reviewed the annotations of the other.

\subsubsection{Implementation details}
The AR baselines (SlowFast~\cite{slowfast} and MViTv2~\cite{mvitv2}) are trained using the SlowFast library~\cite{slowfast}, and some baselines are pre-trained on MECCANO~\cite{MEC}. For all baselines, the configuration yaml files are provided on the dataset repository, such that they can directly be used to reproduce the reported results. The baselines are trained on one Nvidia Tesla v100 GPU.

\begin{table}
  \begin{center}
    {\small{
\begin{tabular}{ll|cc}
\toprule
Model                       & Modality& \begin{tabular}[c]{@{}c@{}}Top-1\\ acc. {[}\%{]}\end{tabular} & \begin{tabular}[c]{@{}c@{}}Top-5\\ acc. {[}\%{]}\end{tabular} \\
\midrule
SlowFast~\cite{slowfast}    & RGB            & \textbf{60.39}    & \textbf{85.21} \\
SlowFast~\cite{slowfast}    & Depth          & 43.20             & 73.98 \\
SlowFast~\cite{slowfast}    & Visible light  & 53.75             & 81.48 \\
SlowFast~\cite{slowfast}    & Stereo images  & 57.72             & 83.03 \\
\midrule
MViTv2~\cite{mvitv2}    & RGB            & \textbf{65.25}    & \textbf{87.93}\\
MViTv2~\cite{mvitv2}    & Depth          & 49.08             & 76.51 \\
MViTv2~\cite{mvitv2}    & Visible light  & 58.59             & 83.50 \\
MViTv2~\cite{mvitv2}    & Stereo images  & 58.86             & 83.55 \\
\bottomrule
\end{tabular}
}}
\end{center}
\caption{AR performance benchmark on IndustReal per modality. All models are pre-trained on Kinetics~\cite{kinetics-400}.}
\label{tab: ar_modalities}
\end{table}

\subsubsection{Quantitative and qualitative analysis}
Figure \ref{fig: ar_conf} shows the normalized confusion matrix for action recognition on the IndustReal test set. Specifically, it shows the performance of the MViTv2~\cite{mvitv2} transformer, pretrained on the Kinetics dataset~\cite{kinetics-400}. It is observed that although the performance is generally adequate, the model confuses actions that are visually similar, such as \textit{take} and \textit{put} the \textit{short\_brace}, and \textit{tighten} and \textit{loosen} the \textit{acorn\_nut}. 

Furthermore, the performance of both the MVit and SlowFast architectures on the RGB, depth, visible light, and stereo image modalities are outlined in~\ref{tab: ar_modalities}. For both architectures, RGB outperforms the other modalities, likely due to the extensive pre-training on the (RGB) Kinetics~\cite{kinetics-400} dataset. Due to hardware limitations on the current HL2 operating system, it is not possible to access the short-throw depth and RGB camera simultaneously, hence the short-throw depth data are not provided. However, the stereo images can be used to generate high-resolution depth maps, if required.

\subsection{Assembly state detection}
\label{sec: ASD}
\subsubsection{Implementation details}
All ASD baselines make use of the YOLOv8-m~\cite{YOLOv8} backbone. The baselines were trained with a learning rate of 5e-4 using the Adam optimizer, warm-up of 0.5~epoch, patience of 5~epochs, and early stopping enabled. Data augmentation is limited to HSV (with default fractions) and image scaling with a +/- gain of 0.2 for all models. For the model trained solely on synthetic data, random occlusions and image mix-up are used as additional data augmentation techniques. The random occlusions are generated using a single rectangle with random color and size, that covers at least 50\% of the bounding box and forces the class label to be background. These occlusions are generated on 33\% of the training images. Image mix-up is performed by merging a synthetic training image with a random image taken from VOC2012~\cite{VOC}, weighing the VOC image with a random factor between 0 and 0.2. 

All baselines are trained on a single Nvidia Tesla v100.

\subsubsection{Quantitative and qualitative analysis}
\begin{figure}[t]
  \centering
  \begin{subfigure}{0.49\linewidth}
    \includegraphics[width=0.99\linewidth]{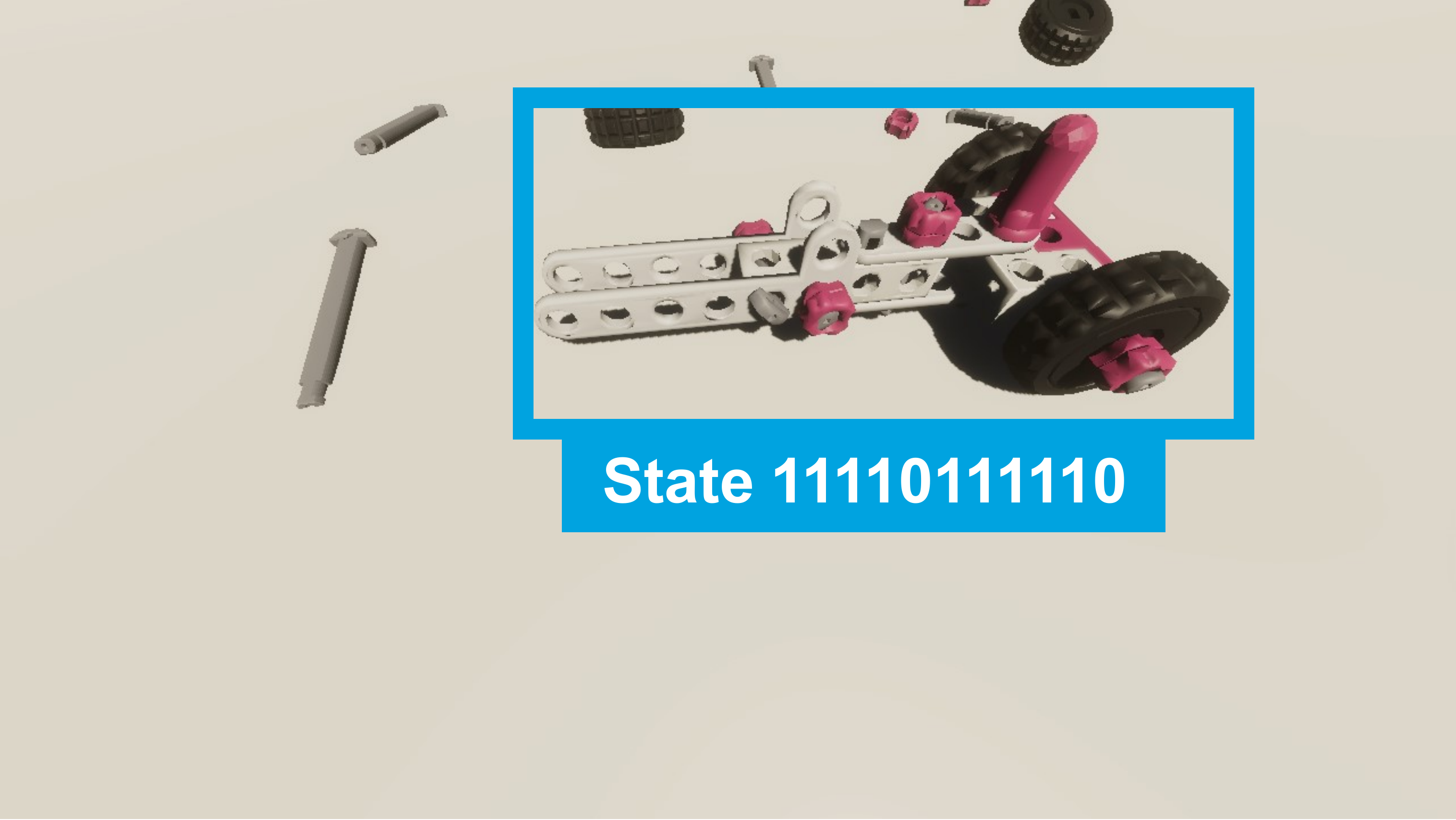}
    \caption{Sample one.}
    \label{fig: synt1}
  \end{subfigure}
  \hfill
  \begin{subfigure}{0.49\linewidth}
    \includegraphics[width=0.99\linewidth]{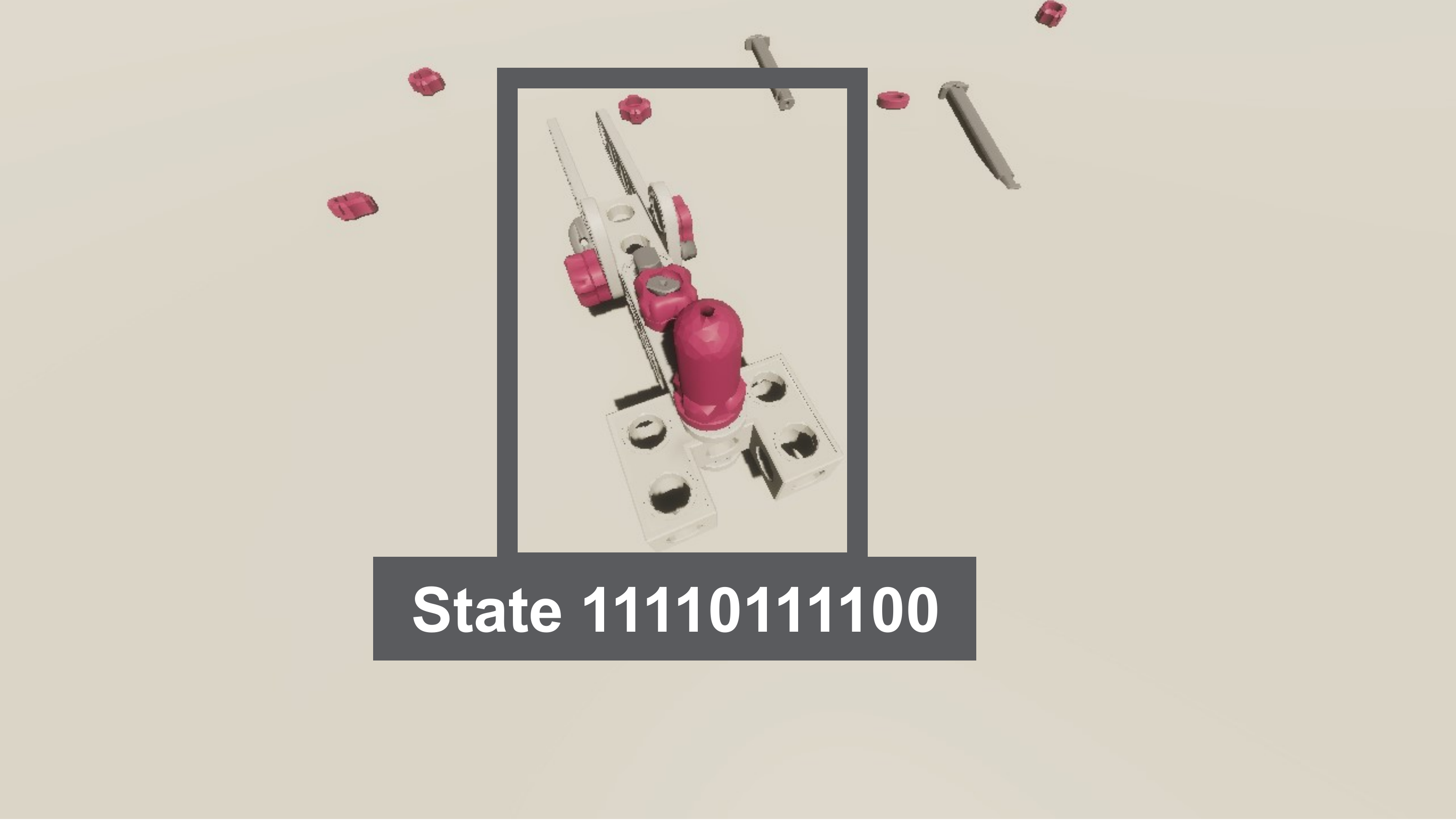}
    \caption{Sample two.}
    \label{fig: synt2}
  \end{subfigure}
  \caption{Samples from the synthetic dataset, generated using the publicly available 3D~models of all parts and Unity Perception \cite{unity_perception}.}
  \label{fig: synt}
\end{figure}

\begin{figure}[t]
\begin{center}
\includegraphics[width=0.99\linewidth]{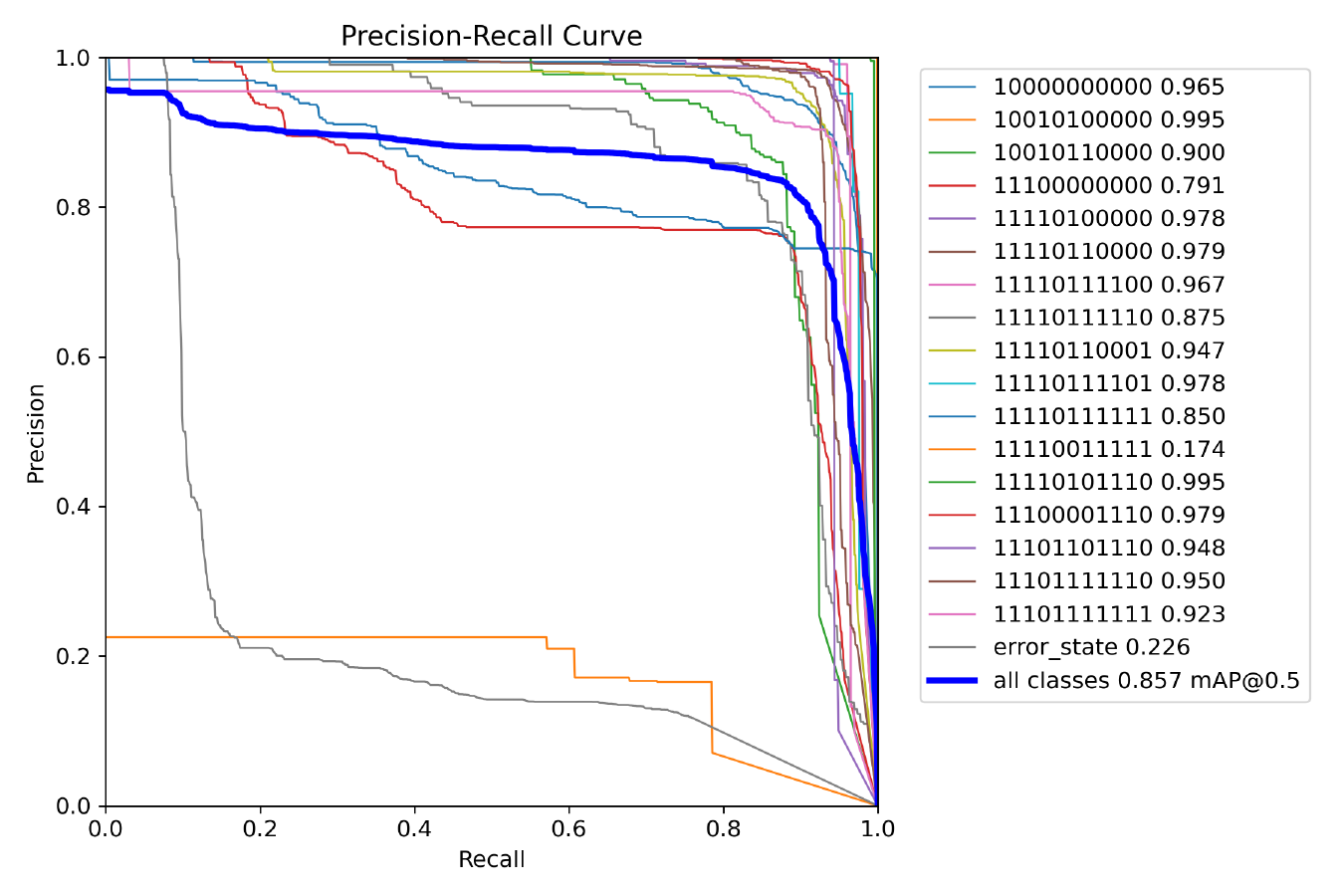}
\end{center}
  \caption{Precision-recall curve on ASD for YOLOv8-m~\cite{YOLOv8}, trained on a combination of real and synthetic data, evaluated on the IndustReal test set.}
\label{fig: PR_curve}
\end{figure}

\begin{figure*}[t]
  \centering
  \begin{subfigure}{0.24\linewidth}
    \includegraphics[width=0.99\linewidth]{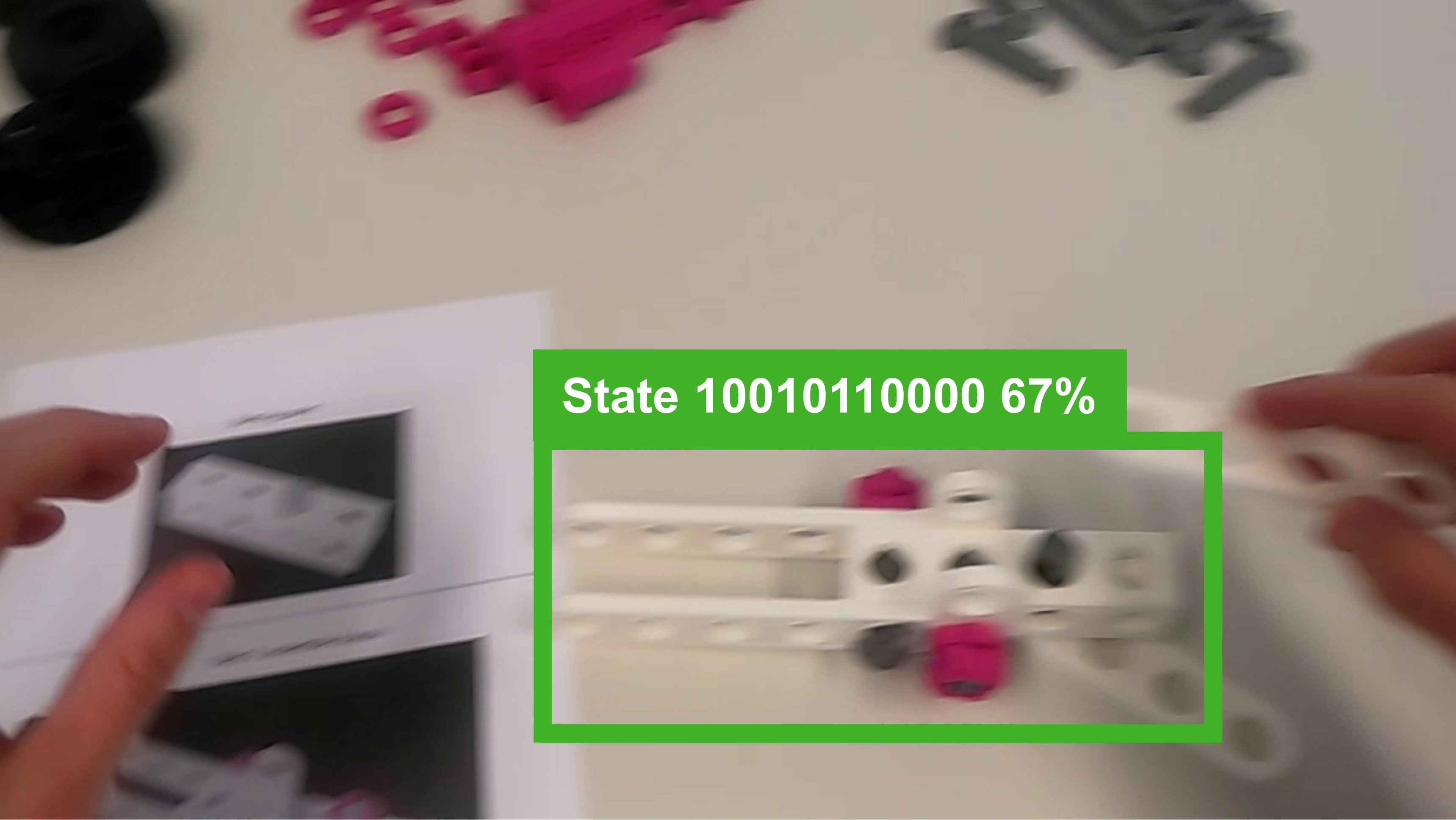}
    \caption{Prediction of the correct class, despite motion blur and a part occluded by the assembly. \newline}
    \label{fig: asd1}
  \end{subfigure}
  \hfill
  \begin{subfigure}{0.24\linewidth}
    \includegraphics[width=0.99\linewidth]{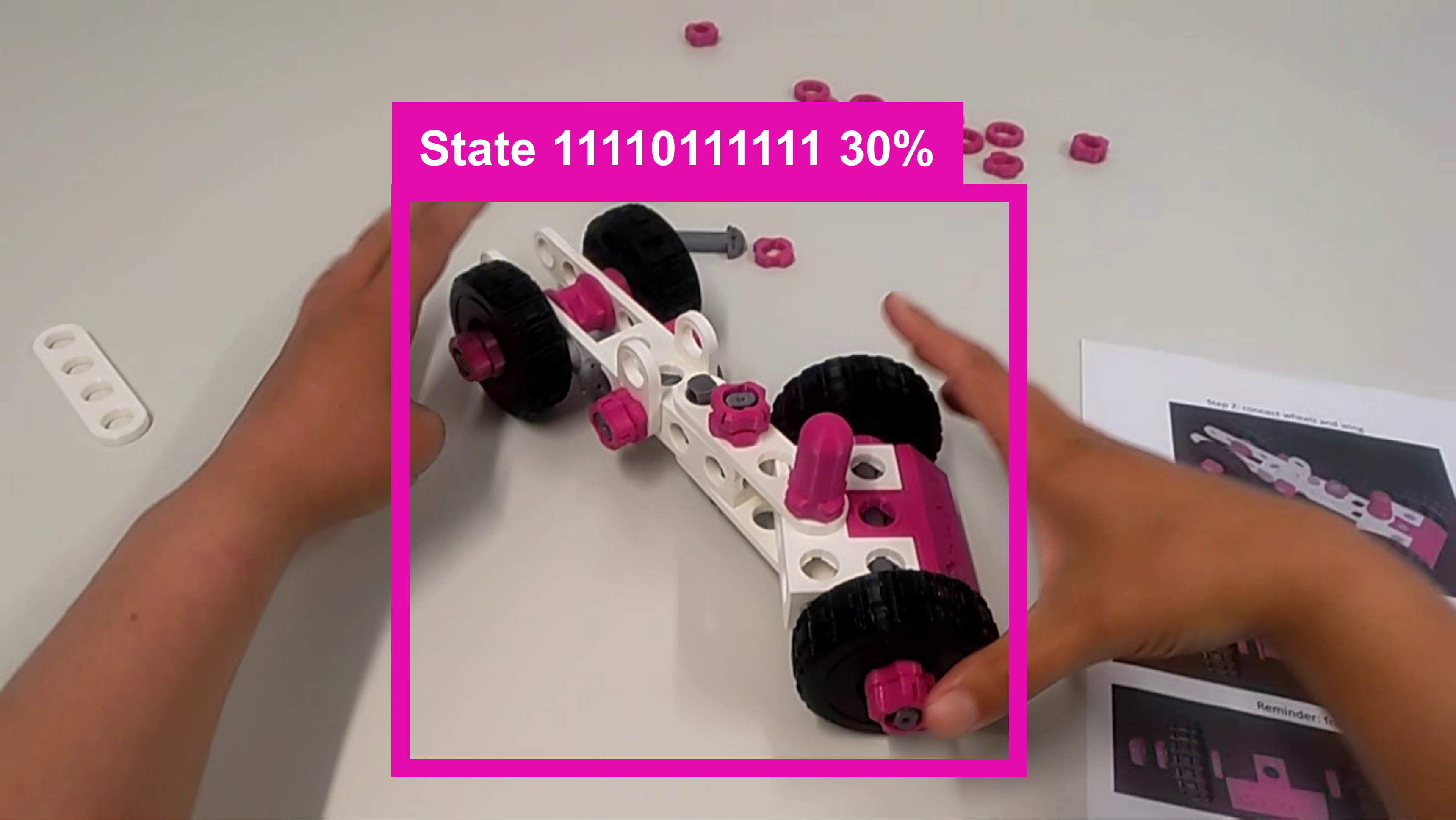}
    \caption{Correct prediction (with low confidence) for final state, despite a clearly misaligned front wheel assembly.}
    \label{fig: asd2}
  \end{subfigure}
  \hfill
  \begin{subfigure}{0.24\linewidth}
    \includegraphics[width=0.99\linewidth]{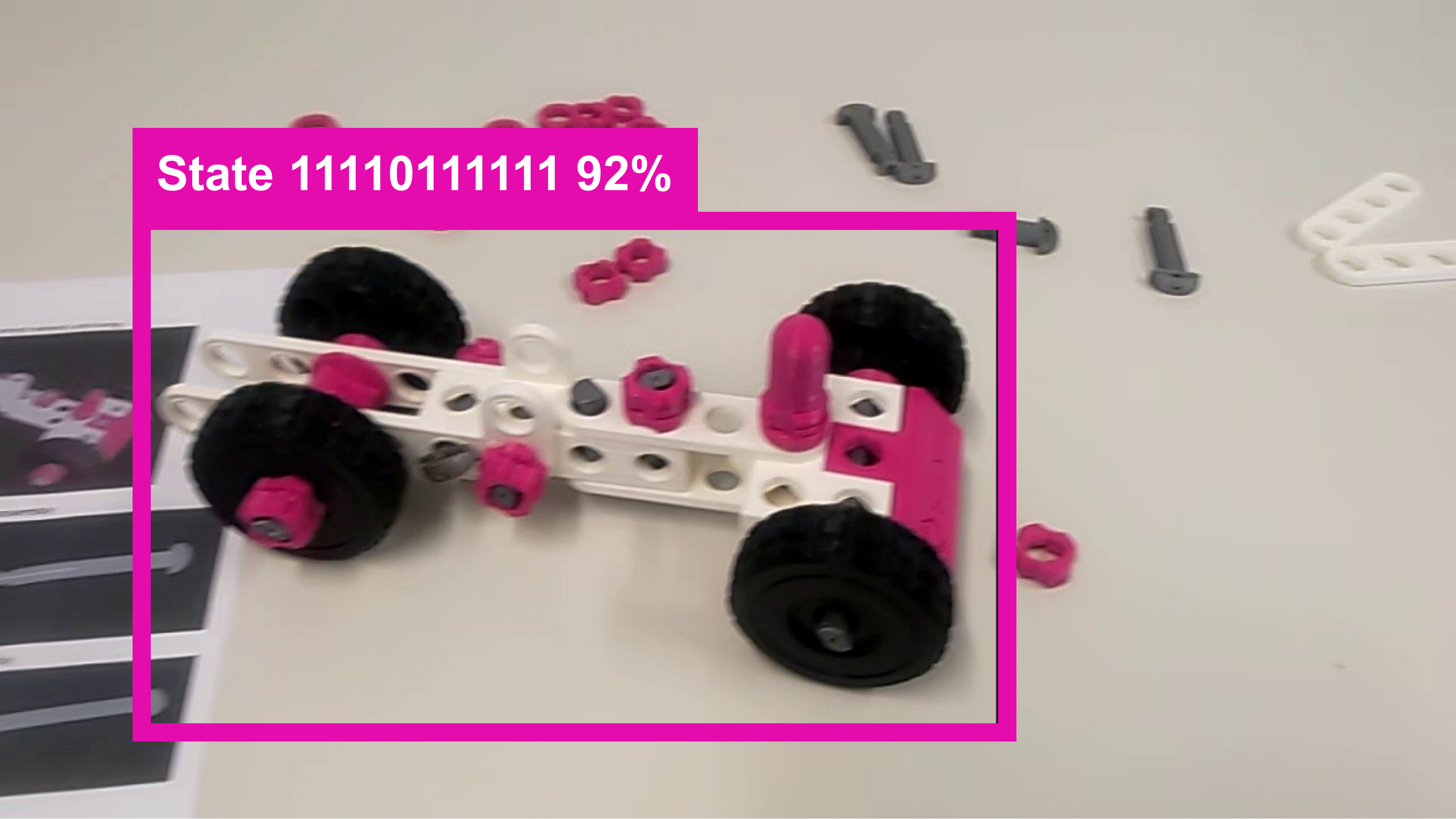}
    \caption{Incorrect prediction (with high confidence) for the completed model, while the washer and nut for the front wheel assy are not yet completed.}
    \label{fig: asd3}
  \end{subfigure}
  \hfill
  \begin{subfigure}{0.24\linewidth}
    \includegraphics[width=0.99\linewidth]{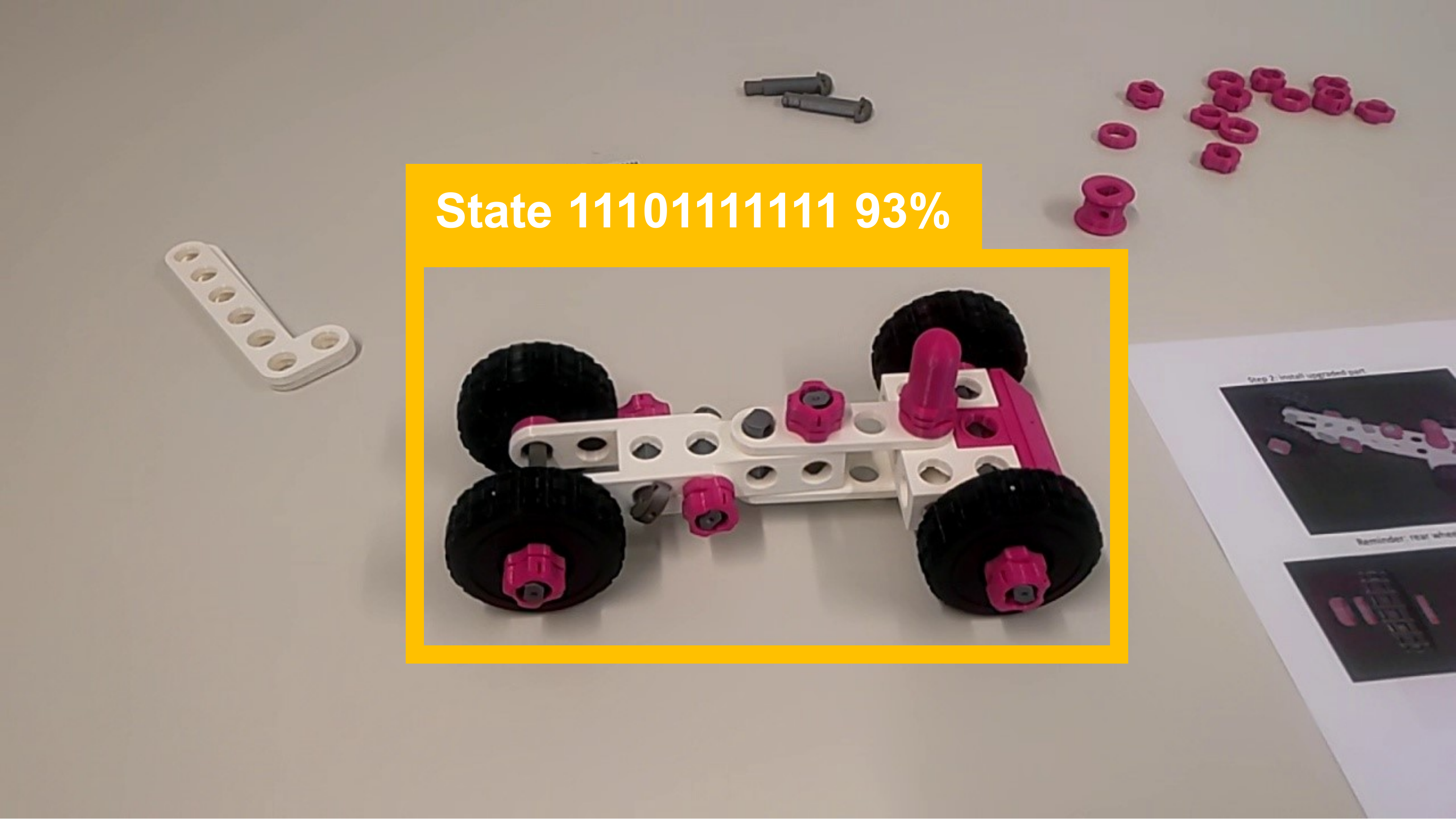}
    \caption{Incorrect prediction (with high confidence) for completed model, as the rear assy is incorrectly assembled due to a missing rear pulley.}
    \label{fig: asd4}
  \end{subfigure}
  \caption{Assembly state detection predictions on the IndustReal test set, using YOLOv8-m \cite{YOLOv8} (B3 in main paper).}
  \label{fig: asd}
\end{figure*}

Two samples from the synthetic dataset constructed with Unity Perception \cite{unity_perception}, used to complement the real-world IndustReal annotations, are demonstrated in Fig.~\ref{fig: synt}. Furthermore, the precision-recall curve of B3 on the test set is shown in Fig.~\ref{fig: PR_curve}. Figure \ref{fig: asd} highlights the prediction of the B3 ASD baseline on the IndustReal test set. The figure highlights correct and incorrect predictions for challenging samples.

\subsection{Procedure step recognition}
\label{sec: PSR}
\subsubsection{Procedure order similarity metric}
\label{sec: POS}
The main paper proposes to measure the quality of a predicted sequence order $\hat{y}$ (e.g., `ACB') by comparing it to a similarity measure with respect to the ground-truth sequence $y$ (e.g., `ABC') using string similarity. Note that this is a simplified denotation of $\hat{y}$, as it does not include the prediction time~$\hat{t}_{\sigma (i)}$ and confidence~$c_{\sigma (i)}$. Approaches on temporal action segmentation use the Levenshtein (Lev) distance\cite{lea2016edit}, normalized over the length of the ground-truth sequence . We propose to make two changes to this metric, which are outlined in more detail here than in the section in the main paper.

The edit distance is defined as the minimum number of edits required to go from a predicted to a ground-truth sequence. For the Lev distance, there are three possible edits to a sequence: insertions, deletions, and substitutions. Insertions and deletions refer to inserting or removing an element of the sequence, e.g. ``ABC\underline{ }'' to ``ABC\underline{D}'' requires the insertion of a ``D''. Substitutions consist of replacing an existing sequence prediction for another, e.g. ``ABC\underline{E}'' to ``ABC\underline{D}'' requires a single substitution of ``E'' to ``D''. The first change to the edit distance proposed in \cite{lea2016edit} is to weight substitutions with a factor of~2, rather than weighting them equally as insertions and deletions, which are weighted by a factor of~1. This essentially eliminates substitutions, as a deletion followed by an insertion is of equal edit distance. The exclusion of substitutions is proposed to prevent the metric from favouring models with many false positives. Models that falsely predict a step when it is not observed, would be penalized less than models that do not predict those false steps.

The second proposed modification is to use the Damerau-Levenshtein (DamLev) \cite{damerau:1964:damlevdistance} edit distance, rather than the Lev distance. The DamLev edit distance allows a fourth edit method, namely transpositions, which are swaps between two subsequent elements in the sequence, \eg ``AB\underline{DC}'' to ``AB\underline{CD}''. Transpositions are included in our proposed procedure order similarity, as it is intuitive to penalize ``A\underline{CB}'' less compared to ``\underline{C}A\underline{B}''. Transpositions, like insertions and deletions, are weighed with a factor of~1.

\begin{table}[]
	\centering
    \def\arraystretch{1.1}
    \caption{Behaviour of the procedure order similarity (POS) metric for various predictions to the ground truth sequence~$y$ ``ABCD'', compared to using the Lev-based similarity proposed in~\cite{lea2016edit}. wDamLev indicates the weighted DamLev as proposed in Section~\ref{sec: POS}, prediction errors are are \underline{underlined}.}
    \label{tab: metric}
    \begin{tabular}{l|cccc}
    \toprule
     $\hat{y}$ & AB\underline{DC} & A\underline{D}C\underline{B} & \underline{D}BC\underline{A} & \underline{  }BCD \\ \midrule
     Edits (Lev)                                        & 2     & 2     & 2     & 1 \\
     Edits (DamLev)~\cite{damerau:1964:damlevdistance}  & 1     & 2     & 2     & 1  \\
     Edits (wDamLev)                                    & 1     & 3     & 4     & 1  \\
     \midrule
     Edit (Lev) \cite{lea2016edit}                      & 0.50  & 0.50  & 0.50 & 0.75\\
     POS (DamLev)                                       & 0.75  & 0.50  & 0.50 & 0.75 \\
     POS (wDamLev)                                      & 0.75  & 0.25  & 0.00 & 0.75 \\
    \bottomrule
    \end{tabular}
\end{table}

\begin{table*}[]
    \centering
    \caption{Demonstration of the behaviour of PSR metrics on some example predictions. A perfect POS, $F_1$ score, or delay individually does not give sufficient information regarding overall performance. The combination of the metrics provides the best insight into model quality.}
    \label{tab: all_metrics}
    \begin{tabular}{l||l|lll}
    \toprule
     & Action $\mid$ observation time & POS & $F_1$ score & Delay $\tau$  \\ \midrule
    Ground truth  & $a_0\mid\SI{5}{\hspace{0.155cm}\second}$,\hspace{0.25cm}$a_1\mid\SI{10}{\second}$,\hspace{0.25cm}$a_2\mid\SI{15}{\second}$,\hspace{0.25cm}$a_3\mid\SI{20}{\second}$   & -- & -- & --  \\ \midrule
    Prediction 1  & $a_0\mid\SI{5}{\hspace{0.155cm}\second}$,\hspace{0.25cm}$a_1\mid\SI{10}{\second}$,\hspace{0.25cm}$a_2\mid\SI{15}{\second}$,\hspace{0.25cm}$a_3\mid\SI{20}{\second}$     & 1.00 & 1.00 & \SI{0.0}{\second}  \\ \midrule
    Prediction 2  & $a_0\mid\SI{5}{\hspace{0.155cm}\second}$,\hspace{0.25cm}$a_1\mid\SI{10}{\second}$,\hspace{0.25cm}$a_3\mid\SI{20}{\second}$,\hspace{0.25cm}$a_2\mid\SI{25}{\second}$    & 0.75 & 1.00 & \SI{2.5}{\second}  \\
    Prediction 3  & $a_0\mid\SI{5}{\hspace{0.155cm}\second}$,\hspace{0.25cm}$a_1\mid\SI{10}{\second}$,\hspace{0.25cm}$a_3\mid\SI{20}{\second}$  & 0.75 & 0.86 & \SI{5.0}{\second}      \\
    Prediction 4  & $a_3\mid\SI{20}{\second}$,\hspace{0.25cm}$a_2\mid\SI{25}{\second}$,\hspace{0.25cm}$a_1\mid\SI{30}{\second}$,\hspace{0.25cm}$a_0\mid\SI{35}{\second}$   & 0.00 & 1.00 & \SI{5.0}{\second}  \\
    Prediction 5  & $a_0\mid\SI{5}{\hspace{0.155cm}\second}$,\hspace{0.25cm}$a_1\mid\SI{5}{\hspace{0.155cm}\second}$,\hspace{0.25cm}$a_2\mid\SI{10}{\second}$,\hspace{0.25cm}$a_3\mid\SI{15}{\second}$  & 1.00 & 0.40 & \SI{0.0}{\second} \\
    \bottomrule
    \end{tabular}
\end{table*}

Similarly to Lea \textit{et al.}~\cite{lea2016edit}, it is proposed to normalize the edit distance. The aforementioned work normalizes with respect to the length of either the ground truth or the prediction, depending on which is longer. We propose to always normalize with respect to the length of the ground truth, as this prevents models with many false positives from being normalized favourably. Combined with the elimination of substitutions, the proposed normalization no longer bounds the normalized distance between~0 and~1. For instance, a ground truth of~(1,~2,~3) and a prediction of~(4,~5,~6) would result in a normalized edit distance of~2 (3~deletions followed by 3~insertions). Therefore, we propose to clip all normalized edit distances which are larger than~1. Since no prediction at all would yield a normalized edit distance of unity, distances larger than that are considered particularly poor predictions. Consequently, the difference between bad and worse predictions is uninteresting and not exploited in this metric. Finally, the clipped, normalized edit distance is subtracted from the unity value. This results in a similarity metric, rather than a distance metric. Thus, the procedure order similarity (POS) between $y$ and $\hat{y}$ is defined by Equation 4 in the main paper. A POS value of unity signifies a perfect match between the prediction and the ground truth ($y=\hat{y}$). The behaviour of the POS metric, as well as the DamLev edit distance, is illustrated in Tab.~\ref{tab: metric}. 

As mentioned in the main paper, looking exclusively at the POS metric is not sufficient for two reasons. Firstly, it does not penalize false positive predictions, if the action is later indeed correctly completed. Secondly, POS does not take time delay into account. For instance, a model could guess the order correctly before any step is completed, and obtain a perfect POS score. For these reasons, two complementary metrics are selected: $F_1$~score and the average delay $\tau$. Table~\ref{tab: all_metrics} demonstrates the behaviour of all three metrics and outlines why the combination of them is essential.

\subsubsection{Implementation details}
The three baselines towards PSR that are evaluated in the paper, are further outlined in this section. Specifically, Algorithm~\ref{alg: one} describes B1, Algorithm~\ref{alg: two} describes B2, and Algorithm~\ref{alg: three} describes B3. B1 consists of the most straight-forward approach, where each detected assembly state that differs from the previously observed state, triggers the completion of all steps required to arrive at the detected state. B2 accumulates the prediction confidences over time, therefore filtering the ASD predictions, resulting in improved POS and $F_1$ score at the cost of an increased delay $\tau$. B3 accumulates the confidences, like B2, and additionally restricts the state transitions to the ones that are expected in the procedure. Only the expected transitions are used to predict completed procedure steps. 

\begin{algorithm}
\caption{PSR baseline 1}
\label{alg: one}
\SetKwInOut{Input}{input}
\SetKwInOut{Output}{output}
\Input{List of ASD predictions $\hat{\mathit{ASD}}$ (in video $X_i$)}
\Output{PSR predictions $\hat{y}$}
$\hat{y} \gets \mathrm{empty}\hspace{0.1cm}\mathrm{list}$\;
$\mathit{ASD}_{curr} \gets \hat{\mathit{ASD}}[0]$\;
$T \gets 0.5$ \Comment*[r]{Conf. threshold}
\For{$f \in \{0 \dots \mathrm{len}(\hat{\mathit{ASD}})\}$}{
    $\mathit{state}, \mathit{conf} \gets \mathit{getHighestPred}(\hat{\mathit{ASD}}[f])$\;
    \If{$\mathit{conf} \geq T$}{
        \If{$\mathit{ASD}_{curr} \neq \mathit{state}$}{
            append $\mathit{state}$ to $\hat{y}$\;
            $\mathit{ASD}_{curr} \gets \mathit{state}$\;
        }
    }
}
\end{algorithm}

\begin{algorithm}
\caption{PSR baseline 2}
\label{alg: two}
\SetKwInOut{Input}{input}
\SetKwInOut{Output}{output}
\Input{List of ASD predictions $\hat{\mathit{ASD}}$ (in video $X_i$)}
\Output{PSR predictions $\hat{y}$}
$\hat{y} \gets \mathrm{empty}\hspace{0.1cm}\mathrm{list}$\;
$\mathit{ASD}_{curr} \gets \hat{\mathit{ASD}}[0]$\;
$\mathit{confs} \gets \mathrm{array}\hspace{0.1cm}\mathrm{with}\hspace{0.1cm}\mathrm{zeros}\hspace{0.1cm}\mathrm{for}\hspace{0.1cm}\mathrm{each}\hspace{0.1cm}\mathrm{component}$\;
$T \gets 8.0$ \Comment*[r]{Conf. threshold}
$\mathit{decay} \gets 0.75$ \Comment*[r]{Confidence decay}
\For{$f \in \{0 \dots \mathrm{len}(\hat{\mathit{ASD}})\}$}{
    $\mathit{state}, \mathit{conf} \gets \mathit{getHighestPred}(\hat{\mathit{ASD}}[f])$\;
    \For{$i \in \{0 \dots \mathrm{len}(\mathit{state})\}$}{
        \eIf{$\mathit{ASD}_{curr,i} \neq \mathit{state}_i$}{
            $\mathit{confs}_i \gets \mathit{confs}_i + \mathit{conf}$\;
            \If{$\mathit{confs}_i \geq T$}{
                append $\mathit{state}_i$ to $\hat{y}$\;
                $\mathit{ASD}_{curr,i} \gets \mathit{state}_i$\;
            }
        }{
            $\mathit{confs}_i = \mathit{confs}_i \cdot \mathit{decay}$\;
        }
    }
}
\end{algorithm}

\begin{algorithm}
\caption{PSR baseline 3}
\label{alg: three}
\SetKwInOut{Input}{input}
\SetKwInOut{Output}{output}
\Input{List of ASD predictions $\hat{\mathit{ASD}}$ (in video $X_i$), descriptive set of procedure information $\mathcal{P}$}
\Output{PSR predictions $\hat{y}$}
$\hat{y} \gets \mathrm{empty}\hspace{0.1cm}\mathrm{list}$\;
$\mathit{ASD}_{curr} \gets \mathrm{initialState}(\mathcal{P})$\;
$\mathit{confs} \gets \mathrm{array}\hspace{0.1cm}\mathrm{with}\hspace{0.1cm}\mathrm{zeros}\hspace{0.1cm}\mathrm{for}\hspace{0.1cm}\mathrm{each}\hspace{0.1cm}\mathrm{component}$\;
$\mathit{s_{exp}} \gets \mathit{findExpectedStates(\mathcal{P})}$\;
$T \gets 8.0$ \Comment*[r]{Conf. threshold}
$\mathit{decay} \gets 0.75$ \Comment*[r]{Confidence decay}
\For{$f \in \{0 \dots \mathrm{len}(\hat{\mathit{ASD}})\}$}{
    $\mathit{state}, \mathit{conf} \gets \mathit{getHighestPred}(\hat{\mathit{ASD}}[f])$\;
    \For{$i \in \{0 \dots \mathrm{len}(\mathit{state})\}$}{
        \eIf{$\mathit{ASD}_{curr,i} \neq \mathit{state}_i$}{
            $\mathit{confs}_i \gets \mathit{confs}_i + \mathit{conf}$\;
            \If{$\mathit{confs}_i \geq T\hspace{0.1cm}\&\hspace{0.1cm}\mathit{ASD}_{curr,i} \in \mathit{s_{exp}}$}{
                append $\mathit{state}_i$ to $\hat{y}$\;
                $\mathit{ASD}_{curr,i} \gets \mathit{state}_i$\;
            }
        }{
            $\mathit{confs}_i = \mathit{confs}_i \cdot \mathit{decay}$\;
        }
    }
}
\end{algorithm}

\begin{figure}[t]
\begin{center}
\includegraphics[width=0.99\linewidth,trim={4cm 0cm 4cm 7cm},clip]{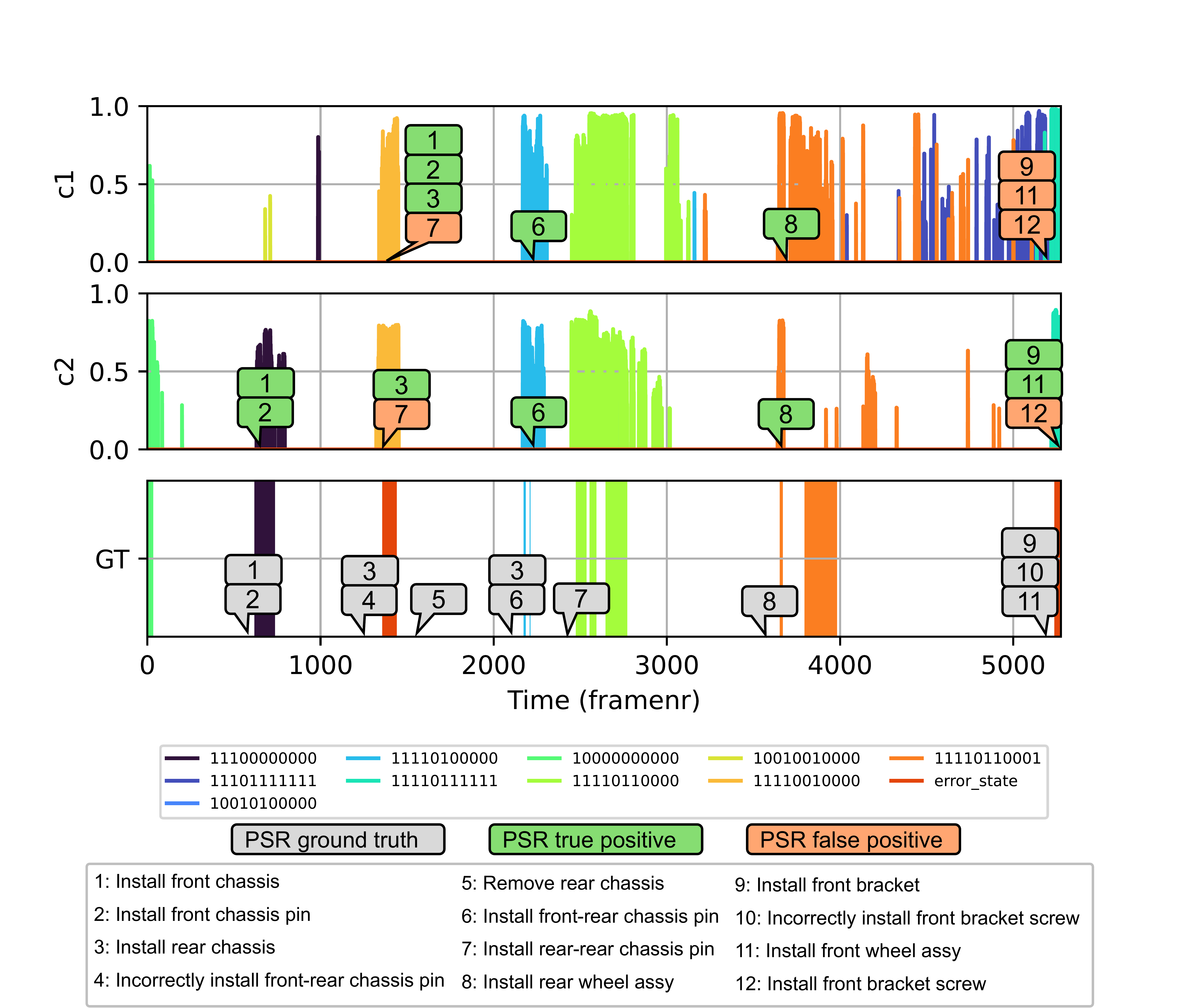}
\end{center}
  \caption{PSR predictions by B3-S (c1) and B3 (c2), together with the accompanied ASD classifications by YOLOv8-m~\cite{YOLOv8}, trained exclusively on synthetic data and on a combination of real and synthetic data, respectively. Predictions are shown for a single video in the IndustReal test set. False and true positive PSR predictions are highlighted.}
\label{fig: psr_qualy}
\end{figure}

\subsubsection{Qualitative analysis}
Figure \ref{fig: psr_qualy} outlines the ASD predictions on a single video in the IndustReal test set for the best scoring ASD approach, as well as the approach trained exclusively on synthetic data. These predictions are used by the PSR baselines and can therefore be used to qualitatively analyze the results from Table 4 in the main paper. It is observed that neither model is able to detect the error state around frame~1300, resulting in false positives for both PSR baselines. The model trained only on synthetic data wrongly classifies the second and last state in the video, explaining the difference in performance between B3 and B3-S outlined in Table 4 in the main paper.

\FloatBarrier

%%%%%%%%% REFERENCES
{\small
\bibliographystyle{ieee_fullname}
\bibliography{IndustReal}
}

\end{document}